\documentclass[letterpaper]{article} % DO NOT CHANGE THIS
\usepackage{aaai24}  % DO NOT CHANGE THIS
\usepackage{times}  % DO NOT CHANGE THIS
\usepackage{helvet}  % DO NOT CHANGE THIS
\usepackage{courier}  % DO NOT CHANGE THIS
\usepackage[hyphens]{url}  % DO NOT CHANGE THIS
\usepackage{graphicx} % DO NOT CHANGE THIS
\usepackage{natbib}  % DO NOT CHANGE THIS AND DO NOT ADD ANY OPTIONS TO IT
\usepackage{caption} % DO NOT CHANGE THIS AND DO NOT ADD ANY OPTIONS TO IT
\usepackage{algorithm}
\usepackage{algorithmic}
\usepackage{enumitem}
\usepackage{amssymb}
\usepackage{amsmath}
\usepackage{graphicx}
\usepackage{subfigure}
\usepackage{newfloat, booktabs}
\usepackage{listings}
\DeclareCaptionStyle{ruled}{labelfont=normalfont,labelsep=colon,strut=off} % DO NOT CHANGE THIS
\floatstyle{ruled}
\newfloat{listing}{tb}{lst}{}
\floatname{listing}{Listing}
\usepackage{hyperref} % for arxiv, hyperlinks are helpful
\usepackage{xcolor}
\hypersetup{
    colorlinks = true,
    linkbordercolor = {white},
    linkcolor={blue},
    urlcolor={blue},
    citecolor= black
}
\nocopyright %no copyright for arxiv version
\title{A Deployed Online Reinforcement Learning Algorithm In An Oral Health Clinical Trial}
\author{
    Anna L. Trella\textsuperscript{\rm 1},
    Kelly W. Zhang\textsuperscript{\rm 2},
    Hinal Jajal\textsuperscript{\rm 1},
    Inbal Nahum-Shani\textsuperscript{\rm 3},
    Vivek Shetty\textsuperscript{\rm 4},
    Finale Doshi-Velez\textsuperscript{\rm 1},
    Susan A. Murphy \textsuperscript{\rm 1}
}
\affiliations {
    \textsuperscript{\rm 1} Department of Computer Science, Harvard University\\
    \textsuperscript{\rm 2} Department of Mathematics, Imperial College London
    \textsuperscript{\rm 3} Institute for Social Research, University of Michigan \\
    \textsuperscript{\rm 4}
    Schools of Dentistry \& Engineering, University of California, Los Angeles \\
    annatrella@g.harvard.edu, hjajal@g.harvard.edu, kelly.zhang@imperial.ac.uk, inbal@umich.edu, vshetty@ucla.edu, finale@seas.harvard.edu, samurphy@g.harvard.edu
}

\usepackage{bibentry}
\usepackage{xcolor}
\usepackage{comment}
\usepackage{color,soul}
\newcommand{\sam}[1]{
  {\color{blue} [SAM: {#1}]}
}

\begin{document}

\maketitle

\begin{abstract}
    Dental disease is a prevalent chronic condition associated with substantial financial burden, personal suffering, and increased risk of systemic diseases. Despite widespread recommendations for twice-daily tooth brushing, adherence to recommended oral self-care behaviors remains sub-optimal due to factors such as forgetfulness and disengagement. To address this, we developed Oralytics, a mHealth intervention system designed to complement clinician-delivered preventative care for marginalized individuals at risk for dental disease. Oralytics incorporates an online reinforcement learning algorithm to determine optimal times to deliver intervention prompts that encourage oral self-care behaviors. We have deployed Oralytics in a registered clinical trial. The  deployment required careful design to manage challenges specific to the clinical trials setting in the U.S. In this paper, we (1) highlight key design decisions of the RL algorithm that address these challenges and (2) conduct a re-sampling analysis to evaluate algorithm design decisions.
    A second phase (randomized control trial) of Oralytics is planned to start in spring 2025.
\end{abstract}

\section{Introduction}
\begin{figure}[!h]
    \centering
    \includegraphics[width=0.4\textwidth]{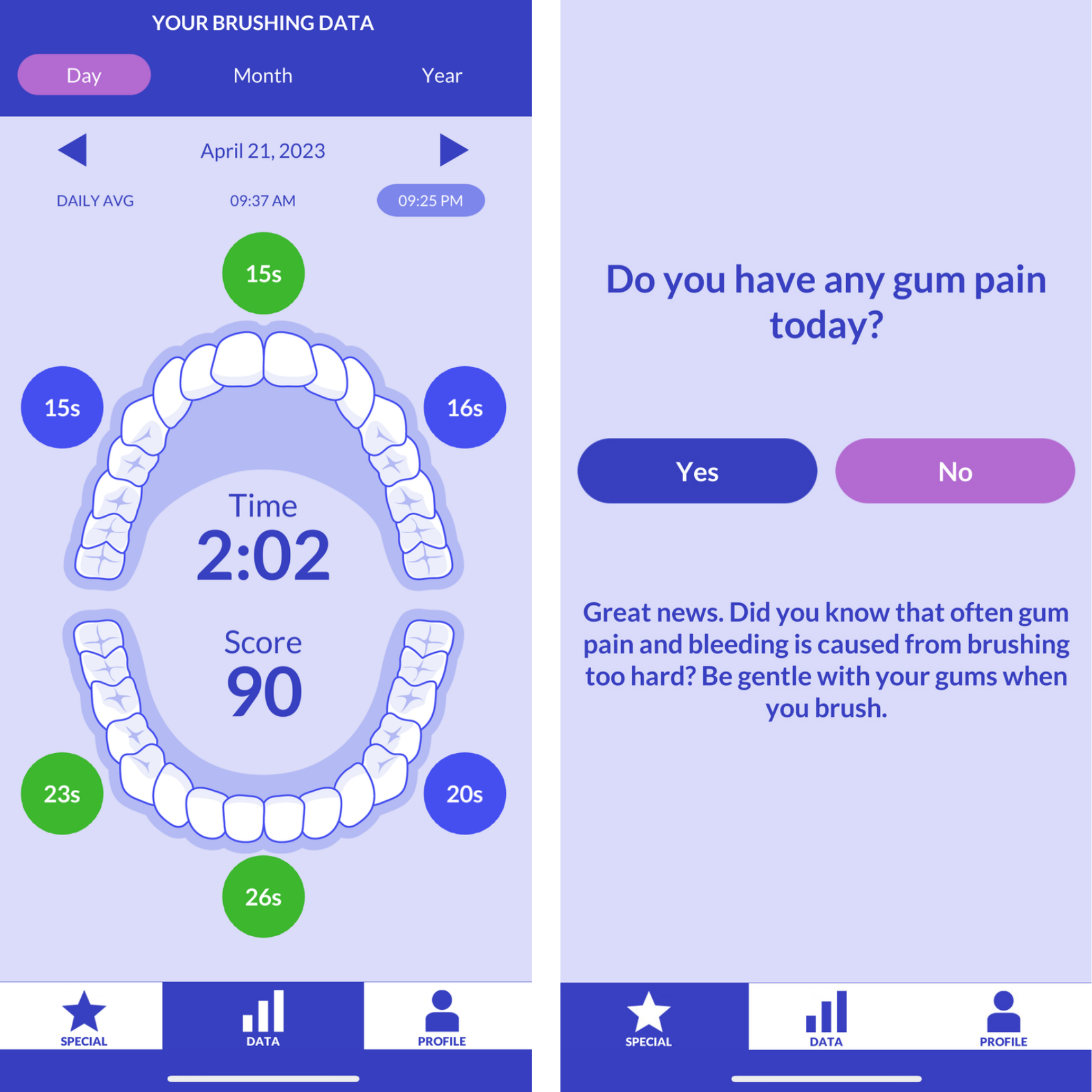}
    \caption{The Oralytics mHealth intervention facilitates high-quality oral self-care behaviors (OSCB) through engagement prompts (e.g., encouraging individuals to monitor their brushing behavior and Q\&A) via the Oralytics app.}
    \label{fig_oralytics_app_ex1}
\end{figure}

% motivating Oralytics
Dental disease is a prevalent chronic condition in the United States with significant preventable morbidity and economic impact \cite{benjamin2010oral}. Beyond its associated pain and substantial treatment costs, dental disease is linked to systemic health complications such as diabetes, cardiovascular disease, respiratory illness, stroke, and adverse birth outcomes. To prevent dental disease, the American Dental Association recommends systematic, twice-a-day tooth brushing for two minutes \cite{ada_recommendation}. 
However, patient adherence to this simple regimen is often compromised by factors such as forgetfulness and lack of motivation \cite{chadwick2011preventive, yaacob2014powered}.

%%% SUSAN'S ORIGINAL PARAGRPH %%%
%\alt{Hello Susan, I have reverted this paragraph back to what you wrote. Could you take a look at the paragraph with mine and Vivek's edits below? It is commented out under ANNA'S EDITS.}
mHealth interventions and tools   can  be leveraged to prompt individuals to engage in high-quality oral self-care behaviors (OSCB) between clinic visits. This work focuses on Oralytics, a mHealth intervention designed to improve OSCB for individuals at risk for dental disease.
% which was deployed in a registered clinical trial \cite{oralytics:clinicaltrial}. 
The intervention involves (i) a Bluetooth-enabled toothbrush to collect sensor data on an individual's brushing quality, and (ii) a smartphone application (app) to deliver treatments, one of which is engagement prompts to encourage individuals to remain engaged in improving their OSCB.
See Figure~\ref{fig_oralytics_app_ex1} for screenshots from the  Oralytics app.
Oralytics includes multiple intervention components one of which is  an  online reinforcement learning (RL) algorithm  which is used to learn, online, a policy specifying when it is most useful to deliver engagement prompts. 
The algorithm should  avoid excessive burden and habituation by only  sending prompts at times they are  likely to be effective.
Before integrating a mHealth intervention
% any health intervention, such as an mHealth intervention, 
into broader healthcare programs, the  effectiveness of the intervention is deployed and tested in a clinical trial. However, the clinical trial setting introduces unique challenges for the design and deployment of online RL algorithms as part of the intervention.

% What should we highlight? Things that people wouldn't think about until they try deploying in real life. Things that are not obvious to a CS audience / things they might not anticipate.
% what makes oralytics special? (1) online algorithm and (2) in a clinical trial that has been in a clinical trial (regulatory constraints, we must pre-register the algorithm that learns online, demands of replicability). (a) you have to pre-register the algorithm and (b) you can't change the algorithm during the trial.

\subsection{Design \& Deployment Challenges in Clinical Trials}
%%% Vivek's suggestion %%%
% Clinical trials operate within a highly regulated environment demanding rigorous adherence to pre-specified protocols. Replicability, essential for verifying results and intervention effectiveness, is paramount. Consequently, algorithmic decisions must be pre-registered and cannot be altered without jeopardizing trial integrity. Even software-related issues during the trial necessitate careful consideration to avoid compromising protocol adherence and replicability.

% \kwz{I feel this next sentence is confusing because we talk about "deployment" challenges, but it says before deployment you need a clinical trial...} 
% \alt{good point Kelly, I think what we mean is before the health intervention is rolled out into a larger health program. Fixed below.}

% \subsubsection{Challenge 1: Transparency and Replicability} 
First, clinical trials, conducted with US National Institutes of Health (NIH) funding, must adhere to the NIH policy on the dissemination of NIH-funded clinical trials \cite{nih_dissemination, clinical_trial_reqs}.
This policy requires pre-registration of the trial in order to enhance  transparency and replicability of trial results (Challenge 1).  The design of the health intervention, including any online algorithms that are components of the intervention, must be pre-registered. 
%Deviations from the pre-registration  can compromise  the transparency and thus harm the ability of others to assess replicability of trial results. 
% the algorithm not functioning as intended due to engineering or networking issues that arise during the trial
Indeed, changing any of the intervention components, including the online algorithm, during the conduct of the trial, makes it difficult for other scientists to know exactly what intervention was implemented and to replicate any results.   
% While one could spontaneously fix issues that arise, 
Thus to enhance transparency and replicability, the online algorithm should be autonomous.  That is, the potential for major ad hoc changes that  alter the pre-registered protocol  should be minimized. 
Second, while the online algorithm learns and updates the policy using incoming data throughout the trial, the algorithm has, in total, a limited amount of data to learn from. 
By design, each individual only receives the mHealth intervention for a limited amount of time. 
Therefore, the RL algorithm only has data on a limited number of decision times for an individual.
This poses a challenge to the RL algorithm's ability to learn based on a small amount of data collected per individual (Challenge 2).

\subsection{Contributions}
In this paper, we discuss how we addressed these deployment challenges in the design of an online RL algorithm – a generalization of a Thompson-sampling contextual bandit (Section~\ref{sec_oralytics_rl_alg_overview}) - as part of the Oralytics intervention to improve OSCB for individuals at risk for dental disease. The RL algorithm (1) learns online from incoming data and (2) makes decisions for individuals in real time as part of the intervention. Recently, the Oralytics intervention was deployed in a registered clinical trial \cite{oralytics:clinicaltrial}. Key contributions of our paper are:
\begin{enumerate}
    \item 
    We highlight key design decisions made for the Oralytics algorithm that deals with deploying an online RL algorithm as part of an intervention in a clinical trial (Section~\ref{sec_deploying_oralytics}).
    \item We conduct a re-sampling analysis\footnote{All code used in this paper can be found in GitHub: \href{https://github.com/StatisticalReinforcementLearningLab/oralytics-post-deployment-analysis/tree/main}{here}} using data collected during the trial to (1) re-evaluate design decisions made and (2) investigate algorithm behavior (Section~\ref{sec_app_payoff}). 
    % \item We offer lessons learned from our experience developing and deploying the Oralytics RL algorithm (Section~\ref{sec_lessons_learned}).
\end{enumerate}

% This paper highlights key challenges and constraints that were managed to ensure the ability to deploy our algorithm in the Oralytics clinical trial \cite{oralytics:clinicaltrial}. 
Further details about the clinical trial and algorithm design decisions can be found in \citet{nahum2024optimizing, trella2024oralytics}.
\section{Related Work}
% this review paper surveyed 41 RCTs of machine learning interventions against  a control: https://jamanetwork.com/journals/jamanetworkopen/fullarticle/2796833

%\subsection
\paragraph{AI in Clinical Trials}
A large body of work exists that incorporates AI algorithms to conduct clinical trials. AI can improve trial execution by automating cohort selection \cite{glicksberg2018automated} and participant eligibility screening \cite{alexander2020evaluation, haddad2021accuracy}. %not deployed
% liu2021evaluating has a method for relaxing some eligibility criteria while maintaining safety 
Prediction algorithms can be used to assist in maintaining retention by identifying participants who are at high risk of dropping out of the trial \cite{pedersen2019predicting, teixeira2022machine}. % not deployed
Recently, generative models have been considered to create digital twins \cite{das2023twin, chandra2024clinicalgan} %, bordukova2024generative} 
of participants to predict participant outcomes or simulate other behaviors. Online algorithms in adaptive trial design \cite{van2019phase, askin2023artificial} can lead to more efficient trials (e.g., time and money saved, fewer participants required) by modifying the experiment design in real-time (e.g., abandoning treatments or redefining sample size). 
The above algorithms are part of the clinical trial design (experimental design)
%trial execution, modify design, or facilitate analyses
while in our setting, the RL algorithm is a component of the  intervention.  
% While these works provide important contributions considering the clinical trials setting, none of these works have been deployed and evaluated in a real trial.
% While these algorithms are subject to the same standards and regulations required in clinical trials, they are not online decision-making algorithms used as part of the intervention.

%\subsection
\paragraph{Online RL Algorithms in mHealth}
% online decision-making algorithms in health
% could not find a registered trial for albers2022addressing but maybe because it was conducted in the netherlands?
% the following have trials registered with clinicaltrials.gov: figueroa2021adaptive, forman2023using, ghosh2024rebandit, yom2017encouraging, lauffenburger2021reinforcement, lauffenburger2024impact, DBLP:journals/corr/abs-1909-03539, piette2022patient
% \alt{Idea:
% \begin{itemize}
%     \item Many online RL algorithms have been included in mHealth interventions. Cite the IAAI paper adaptive bandit experiments and others. There are other deployment of online RL algorithms in mHealth interventions that were not formally in registered clinical trials.
%     \item These interventions were deployed in a registered clinical trial.
%     \item We focus on addressing the transparency and replicability constraints imposed on the RL algorithm in a clinical trial.
% \end{itemize}
% }

Many online RL algorithms have been included in mHealth interventions deployed in a clinical trial. For example, online RL was used to optimize the delivery of prompts to encourage physical activity \cite{yom2017encouraging, DBLP:journals/corr/abs-1909-03539, figueroa2021adaptive}, 
% \sam{I am not sure if all of these cites are about "online" RL--please check...}
% hi Susan, I got these cites from the deployed online RL folder and double-checked that they were indeed online
manage weight loss \cite{forman2023using}, improve medical adherence \cite{lauffenburger2024impact}, %lauffenburger2021reinforcement
assist with pain management \cite{piette2022patient}, reduce cannabis use amongst emerging adults \cite{ghosh2024rebandit}, and help people quit smoking \cite{albers2022addressing}. 
There are also deployments of online RL in mHealth settings that are not formally registered clinical trials \cite{zhou2018personalizing, kumar2024using}. 
% \sam{what about Milind Tambe's paper??  see cite in latex doc} \alt{Milind Tambe's paper is not online RL. They learn a policy offline and deploy it. They say: ``In contrast, we focus on learning RMAB parameters using clustered historic benefciary data". Should I still include?}
%Aditya Mate*, Lovish Madaan*, Aparna Taneja, Neha Madhiwalla, Shresth Verma, Gargi Singh, Aparna Hegde, Pradeep Varakantham, and Milind Tambe. 2/2022. “Field Study in Deploying Restless Multi-Armed Bandits: Assisting Non-Profits in Improving Maternal and Child Health.” In AAAI Conference on Artificial Intelligence. Vancouver, Canada.
Many of these papers focus on algorithm design before deployment.  
Some authors \cite{kumar2024using}, compare outcomes between groups of individuals where each group is assigned a different algorithm or policy.   %compare the  our work additionally performs analyses after deployment to re-evaluate algorithm design decisions.
%which can improve future deployment and applications. 
% This re-evaluation is critical as it can improve future deployments and applications.
%In contrast to \cite{kumar2024using} we  re-evaluate design decisions by deploying online %and testing
%multiple variants of the algorithm \cite{kumar2024using}, this was not the case in Oralytics. 
Here we use a different analysis to inform further design decisions. Our analysis focuses on learning across time by a single online RL algorithm.  
\section{Preliminaries}

\begin{table}[t]
    \centering
        \begin{tabular}{c|c}
            % \hline
            % Header 1 & Header 2 \\
            \hline
            Trial Start & September 2023 \\
            Trial End & July 2024 \\
            Num. Participants & 79 \\
            Recruitment Rate & Around 5 per 2 weeks \\
            Num. of Days Participant in Trial & 70 \\
            Num. Decision Times Per Day & 2 \\
            \hline
        \end{tabular}
    \caption{Oralytics Clinical Trial Facts
    % \sam{usually you list the total number of participants and then in the text, section 3.1 explain why you are restricting analyses to a subset.  I suggest you do that here.}
    }
    \label{tab_oralytics_facts}
\end{table}

\subsection{Oralytics Clinical Trial}
The Oralytics clinical trial (Table~\ref{tab_oralytics_facts}) enrolled participants recruited from UCLA dental clinics in Los Angeles\footnote{
The study protocol and consent procedures have been approved by the University of California, Los Angeles Institutional Review Board (IRB\#21–001471) and the trial was registered on ClinicalTrials.gov (NCT05624489).}. Participants were recruited incrementally at about 5 participants every 2 weeks.
All participants received an electric toothbrush with WiFi and Bluetooth connectivity and integrated sensors. Additionally, they were instructed to download the Oralytics app on their smartphones. 
%As discussed earlier, the Oralytics online RL algorithm
%, designed to improve the proximal outcome of brushing quality, 
%governed the delivery of  engagement prompts  via push notifications from the Oralytics app.
The RL algorithm dynamically decided whether to deliver an engagement  prompt for each participant twice daily, with delivery within an hour preceding self-reported morning and evening brushing times.
%These engagement prompts included content such as motivational messages to encourage individuals to access the Oralytics app for brushing feedback and information about oral health.
% tailored to participants,  \sam{no tailored content} %brushing feedback \sam{I don't think the prompts included brushing feedback--what words does Billie use in paper--we should use those words?}
% \alt{Hi Susan, in the protocol paper, Billie said ``the prompts contain information about oral health and encourage the individual to access the Oralytics app for brushing feedback"}
The clinical trial began in September 2023 and was completed in July 2024. A total of 79 participants were enrolled over approximately 20 weeks, with each participant contributing data for 70 days. 
However, due to an engineering issue, data for 7 out of the 79 participants was incorrectly saved and thus their data is unviable. Therefore, we restrict our analyses (in Section~\ref{sec_app_payoff}) to data from the 72 unaffected participants. For further details concerning the trial design, see \citet{oralytics:clinicaltrial} and \citet{ nahum2024optimizing}.

\subsection{Online Reinforcement Learning}
Here we consider a setting involving sequential decision-making for $N$ participants, each with $T$ decision times.
Let subscript $i \in [1:N]$ denote the participant and subscript $ t\in [1:T]$ denote the decision time.  $S_{i, t}$ denotes the current state of the participant.  At each decision time $t$, the algorithm selects action $A_{i, t}$ after observing $S_{i, t}$, based on its policy $\pi_{\theta}(s)$ which is a function, parameterized by $\theta$, that takes in input state $s$. After executing action $A_{i, t}$, the algorithm receives a reward $R_{i, t}$. In contrast to batch RL, where policy parameters are learned using previous batch data and fixed for all $t \in [1: T]$, online RL learns the policy parameters with incoming data. At each update time $\tau$, the algorithm updates parameters $\theta$ using the entire history of state, action, and reward tuples observed thus far $\mathcal{H}_\tau$. The goal of the algorithm is to maximize the average reward across all participants and decision times, $\mathbb{E} \big[ \frac{1}{N\cdot T}\sum_{i = 1}^N \sum_{t = 1}^T R_{i, t} \big]$.

\subsection{Oralytics RL Algorithm}
\label{sec_oralytics_rl_alg_overview}
The Oralytics RL algorithm is a generalization of a Thompson-Sampling contextual bandit algorithm \cite{russo2018tutorial}. The algorithm makes decisions at each of the $T = 140$ total decision times ($2$ every day over $70$ days) on each participant.
The algorithm state (Table~\ref{tab_state}) includes current context information about the participant collected via the toothbrush and app (e.g., participant OSCB over the past week and prior day app engagement).
The RL algorithm makes decisions regarding whether or not to deliver an engagement prompt to  each participant twice daily, one hour before a participant's self-reported usual morning and evening brushing times. Thus the action space is binary, with $A_{i, t}=1$ denoting delivery of the prompt and $A_{i, t}=0$, otherwise. 

The reward, $R_{i, t}$, is constructed based on the proximal health outcome OSCB, $Q_{i, t}$, and a tuned approximation to the effects of actions on future states and rewards. This reward design allows a contextual bandit algorithm to approximate an RL algorithm that models the environment as a Markov decision process.
See \citet{trella2023reward} for more details on the reward designed for Oralytics. 

% \kwz{I would also suggest combining/reorganizing content in 3.2 and 3.3 since there is a lot of repetition.} 
As part of the policy, contextual bandit algorithms use a model of the mean reward given state $s$ and action $a$, parameterized by $\theta$: $r_{\theta}(s, a)$. We refer to this as the reward model. While one could learn and use a reward model per participant $i$, in Oralytics, we ran a full-pooling algorithm (Section~\ref{sec_pooling}) that learns and uses a single reward model shared between all participants in the trial instead. In Oralytics, the reward model $r_{\theta}(s, a)$ is a linear regression model as in \citet{DBLP:journals/corr/abs-1909-03539} (See Appendix~\ref{app_reward_model}). The Thompson-Sampling algorithm is Bayesian and thus the algorithm has a prior distribution $\theta \sim \mathcal{N}(\mu^{\text{prior}}, \Sigma^{\text{prior}})$ assigned to parameter $\theta$. See Appendix~\ref{app_prior} for the prior designed for Oralytics.

%%%%% UPDATE %%%%%
The RL algorithm updates the posterior distribution for parameter $\theta$ once a week on Sunday morning using all participants' data observed up to that time; denote these weekly update times by $\tau$. 
Let $n_\tau$ be the number of participants that have started the trial before update time $\tau$, and $t(i, \tau)$ be a function that takes in participant $i$ and current update time $\tau$ and outputs the last decision time for that participant. Then to update posterior parameters $\mu^{\text{post}}_\tau, \Sigma^{\text{post}}_\tau$, we use the history $\mathcal{H}_{\tau} := \{(S_{i, t'}, A_{i, t'}, R_{i, t'})\}_{i = 1, t' = 1}^{n_\tau, t(i, \tau)}$. 
Thus the RL algorithm is a full-pooling algorithm that pools observed data, $\mathcal{H}_{\tau}$ from all participants to update posterior parameters $\mu^{\text{post}}_\tau, \Sigma^{\text{post}}_\tau
$ of $\theta$.
% at update time, $\tau$.
% \kwz{I think incremental recruitment hasn't been mentioned yet(?).} 
% \alt{thanks for pointing out Kelly! I added incremental recruitment details to the Oralytics clinical trial section earlier.}
Notice that due to incremental recruitment of trial participants, at a particular update time $\tau$, not every participant will be on the same decision time index $t$ and the history will not necessarily involve all $N$ participants' data.

%%%%% ACTION-SELECTION %%%%%

% \sam{this section could be 1/2 as long...} 
%\kwz{I think below might be helpful use the $A_{i,t}$ notation, which is used above to introduce that actions are binary.} 
% \alt{Anna tried to condense and make this section more clear.}

To select actions, the RL algorithm uses the latest reward model to model the \textit{advantage}, or the difference in expected rewards, of action $1$ over action $0$ for a given state $s$. 
Since the reward model for Oralytics is linear, the model of the advantage is also linear:

\begin{align}
\label{eqn_adv_model}
    r_{\theta}(s, a=1) - r_{\theta}(s, a=0) = f(s)^\top \beta
\end{align}

$f(s)$ denotes the features used in the algorithm’s model for the advantage (See Table~\ref{tab_state}), and $\beta$ is the subset of parameters of $\theta$ corresponding to the advantage. 
% where $f(s)^\top \beta$ is the model of the advantage of action $1$ over action $0$,
For convenience, let $\tau = \tau(i, t)$ be the last update time corresponding to the current reward model used for participant $i$ at decision time $t$. The RL algorithm micro-randomizes actions using $\mathbb{P}(f(s)^\top \beta > 0  | s = S_{i, t}, \mathcal{H}_{\tau})$ and therefore forms action-selection probability $\pi_{i, t}$:
\begin{align}
\label{eqn_action_selection_prob}
    \pi_{i,t} := \mathbb{E}_{\beta \sim \mathcal{N}(\mu^{\beta}_\tau, \Sigma^{\beta}_\tau )}\left[\rho(f(s)^\top \beta) \big| s = S_{i, t}, \mathcal{H}_{\tau} \right]
\end{align}
where $\mu^{\beta}_{\tau}$ and $\Sigma^{\beta}_{\tau}$ are the sub-vector and sub-matrix of $\mu^{\text{post}}_\tau$ and $\Sigma^{\text{post}}_\tau$ corresponding to advantage parameter $\beta$.
% where $\mathcal{H}_{\tau(i, t)}$ denotes the history of state, action, and rewards observed up to update time $\tau(i, t)$. 
Notice that while classical posterior sampling uses an indicator function for $\rho$, the Oralytics RL algorithm instead uses a generalized logistic function for $\rho$ to ensure that policies formed by the algorithm concentrate and enhance the replicability of the algorithm \cite{zhang2024replicable}. %zhang2020inference, zhang2022statistical, 

Finally, the RL algorithm samples $A_{i, t}$ from a Bernoulli distribution with success probability $\pi_{i, t}$:
\begin{align}
    A_{i, t} \mid \pi_{i, t} \sim \text{Bern}(\pi_{i, t})
\end{align}
%\kwz{For a mathy reader, above should be conditional statement. $A_{i, t} \mid S_{i,t}, H_{t-1} \sim \text{Bern}(\pi_{i, t})$}
%\alt{Can I do the above and just condition on $\pi_{i, t})$? because that's a value we calculate.} \kwz{Yes!}

%%%%% UPDATES %%%%%$
% \sam{Anna, following text needs to be fixed.  Above $t$ indexed user's decision times during user's 70 day intervention.  In the following $t$ indexes decision times from when the first user started the intervention to when the last user finished their 70 days.  I suggest to use something like $\iota$ and tell reader this.  Note  there are a total of $N\cdot T$ decision times in the trial ($N$ participants, each with $T$ decisions).  Let  $n_\iota$ denote the number of participants present in the trial at decision time $\iota\in[1:N\cdot T$.} 
% \alt{confusion and discussion resolved through email.}
%\kwz{I feel again that the updating was introduced in first paragraph in section 3.2 and now reintroduced with more detail. Could consolidate these...} 

\section{Deploying Oralytics}
\label{sec_deploying_oralytics}

\begin{figure*}[t]
    \centering    \includegraphics[trim=0 7cm 0 0, clip, width=1\textwidth]{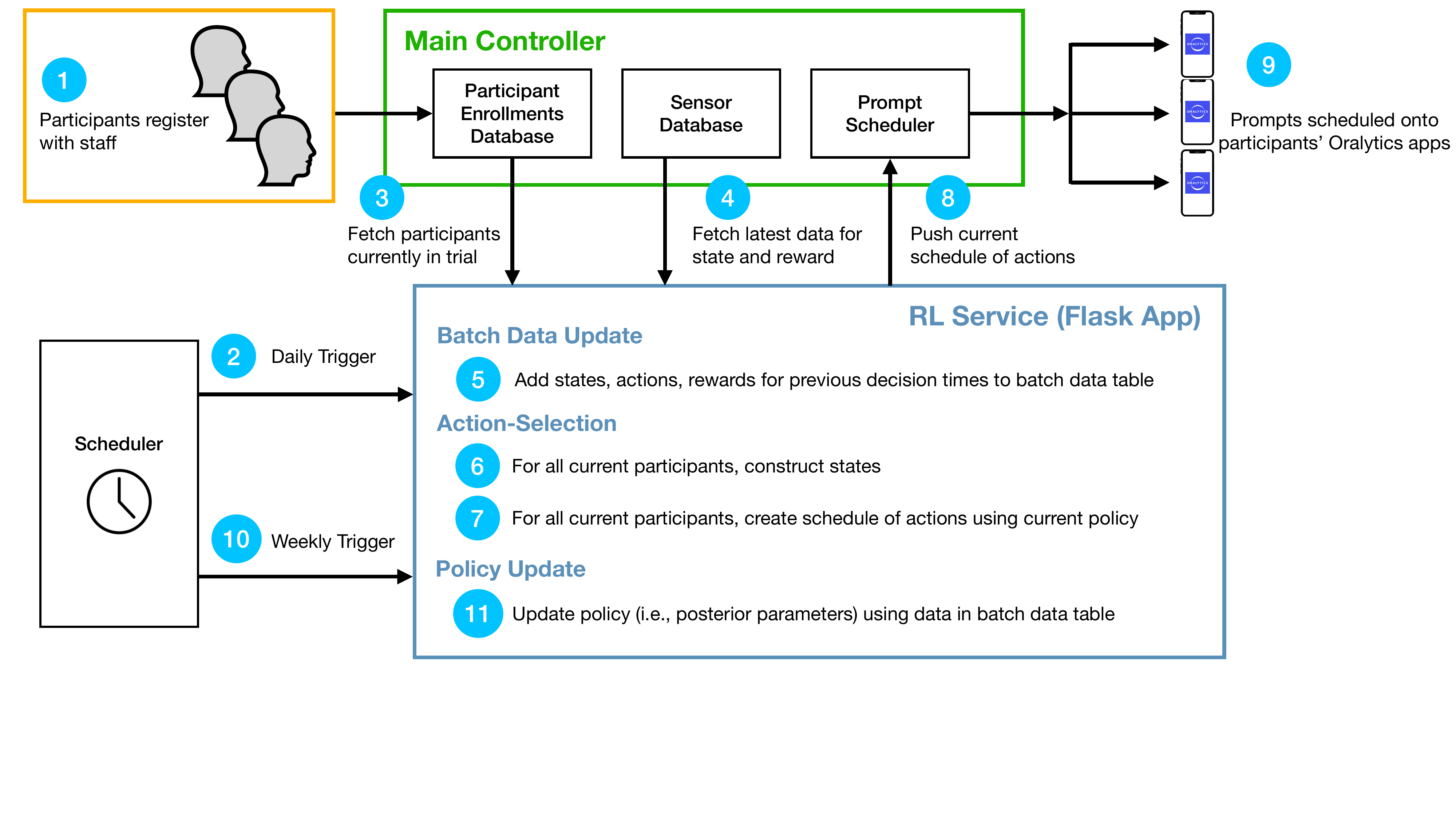}  % Adjust width as needed
    \caption{Oralytics End-to-End Pipeline.}
    \label{fig_system_architecture}
\end{figure*}

\subsection{Oralytics Pipeline}
\paragraph{Software Components}
%To function properly, the Oralytics clinical trial involved interactions with
Multiple software components form the Oralytics software service. These components are (1) the main controller, (2) the Oralytics app, and (3) the RL service. The \emph{main controller} is the central coordinator of the Oralytics software system that handles the logic for (a) enrolling participants, (b) pulling and formatting sensor data (i.e., brushing and app analytics data), and (c) communicating with the mobile app to schedule prompts for every participant. The \textit{Oralytics app} is downloaded onto each participant's smartphone at the start of the trial. The app is responsible for (a) obtaining prompt schedules for the participant and scheduling them in the smartphone's internal notification system and (b) providing app analytics data to the main controller. The \textit{RL service} is the software service supporting the RL algorithm to function properly and interact with the main controller. The RL service executes three main processes: (1) batch data update, (2) action selection, and (3) policy update. 

The main controller and RL service were deployed on infrastructure hosted on Amazon Web Services (AWS). Specifically, the RL service was wrapped as an application using Flask. A daily scheduler job first triggered the batch data update procedure and then the action-selection procedure and a weekly scheduler job triggered the policy update procedure. The Oralytics app was developed for both Android and iOS smartphones.

\paragraph{End-to-End Pipeline Description}
We now describe interactions between clinical staff with components of the Oralytics software system  and between software components (See Figure~\ref{fig_system_architecture}). The Oralytics clinical trial staff recruits and registers participants (Step 1). The registration process consists of the participant downloading the Oralytics app and staff verifying that the participant had at least one successful brushing session from the toothbrush. Successfully registered participants are then entered into the participant enrollment database maintained by the main controller. The main controller maintains this database to track participants entering and completing the trial (i.e.,  at 70 days).

Every morning, a daily scheduler job first triggers the batch data update process and then the action-selection process (Step 2). The RL service begins by fetching the list of participants currently in the trial (Step 3) and the latest sensor data (i.e., brushing and app analytics data) for current participants (Step 4) from the main controller. 
Notice that this data contains rewards to be associated with previous decision times as well as current state information.
Rewards are matched with the correct state and action and these state, action, and reward tuples corresponding to \textit{previous} decision times are added to the RL service's internal batch data table (Step 5).  During the action-selection process, the RL service first uses the latest sensor data to form states for all current participants (Step 6). Then, the RL service uses these states and the current policy to create a new schedule of actions for all current participants (Step 7). 
These states and actions are saved to the RL internal database to be added to the batch data table during Step 5, the next morning.
All new schedules of actions are pushed to the main controller and processed to be fetched (Step 8). When a participant opens their Oralytics app, the app fetches the new prompt schedule from the main controller and schedules prompts as notification messages in the smartphone's internal notification system (Step 9).

Every Sunday morning, a weekly scheduler job triggers the policy update process (Step 10). During this process, the RL system takes all data points (i.e., state, action, and reward tuples) in the batch data table and updates the policy (Step 11). Recall that the Oralytics RL algorithm is a Thompson sampling algorithm which means policy updates involve updating the posterior distribution of the reward model parameters (Section~\ref{sec_oralytics_rl_alg_overview}). The newly updated posterior distribution for the parameters is used to select treatments for all participants and all decision times for that week until the next update time.

Every morning, the Oralytics pipeline (Steps 6-8) produces a full 70-day schedule of treatment actions  for each participant starting at the current decision time (as opposed to a single action for the current decision time). The schedule of actions is a \textit{key design decision for the Oralytics system} that enhances the transparency and replicability of the trial (Challenge 1). 
Specifically, this design decision mitigates networking or engineering issues if: (1) a new schedule of actions fails to be constructed or (2) a participant does not obtain the most recent schedule of actions. We further see the impact of this design decision during the trial in Section~\ref{sec_fallback_impact}.

\subsection{Design Decisions To Enhance Autonomy and Thus Replicability}
\label{sec_fallback}
A primary challenge in our setting is the high standard for replicability and as a result the algorithm, and its components, should be autonomous (Challenge 1). However, unintended engineering or networking issues could arise during the trial. These issues could cause the intended RL system to function incorrectly compromising: (1) participant experience and (2) the quality of data for post-trial analyses. 
% \subsubsection{Fallback Methods}

One way Oralytics dealt with this constraint is by implementing \textit{fallback methods}. Fallback methods are pre-specified backup procedures, for action selection or updating, which are executed when an issue occurs. Fallback methods are part of a larger automated monitoring system \cite{trella2024monitoring} 
%\alt{Cite Susobhan and Anna JAMIA paper when it comes out} \sam{will not come out in time} \alt{sad :(} %trella2024effectivemonitoringonlinedecisionmaking
that detects and addresses issues impacting or caused by the RL algorithm in real-time. Oralytics employed the following fallback methods: 
\begin{enumerate}[label=(\roman*)]
    \item \label{oralytics_fb1} 
    if any issues arose with a participant not obtaining the most recent schedule of actions, then the action for the current decision time will default to the action for that time from the last schedule pushed to the participant's app.
    \item \label{oralytics_fb2} if any issues arose with constructing the schedule of actions, then the RL service forms a schedule of actions where each action is selected with probability 0.5 (i.e., does not use the policy nor state to select action).
    \item \label{oralytics_fb3} for updating, if issues arise (e.g., data is malformed or unavailable), then the algorithm stores the data point, but does not add that data point to the batch data used to update parameters.
\end{enumerate}

\subsection{Design Decisions Dealing with Limited Decision Times Per Individual}
\label{sec_pooling}
% \sam{revisit description of challenges and harmonize if needed.}
% \alt{harmonized! Challenge 2 is that because there's limited decision times per individual, therefore we have sparse data or very few data points to learn from on that individual.}
Each participant is in the Oralytics trial for a total of $140$ decision times, which results in a small amount of data collected per participant. Nonetheless, the RL algorithm needs to learn and select quality actions based on data from a limited number of decision times per participant (Challenge 2).

A design decision to deal with limited data is \textit{full-pooling}. Pooling refers to clustering participants and pooling all data within a cluster to update the cluster's shared policy parameters. Full pooling refers to pooling all $N$ participants' data together to learn a single shared policy. 
% Pooling algorithms have shown good empirical performance when participants within a cluster are similar \cite{zhu2018group,tomkins2021intelligentpooling}. 
Although participants are likely to be heterogeneous (reward functions are likely different), we chose a full-pooling algorithm like in \citet{yom2017encouraging,figueroa2021adaptive,piette2022patient} to trade off bias and variance in the high-noise environment of Oralytics. These pooling algorithms can reduce noise and speed up learning.    

We finalized the full-pooling decision after conducting experiments comparing no pooling (i.e., one policy per participant that only uses that participant's data to update) and full pooling. We expected the no-pooling algorithm to learn a more personalized policy for each participant later in the trial if there were enough decision times, but the algorithm is unlikely to perform well when there is little data for that participant. Full pooling may learn well for a participant's earlier decision times because it can take advantage of other participants' data, but may not personalize as well as a no-pooling algorithm for later decision times, especially if participants are heterogeneous. In extensive experiments, using simulation environments based on data from prior studies,  we found that full-pooling algorithms achieved higher average OSCB  than no-pooling algorithms across all variants of the simulation environment (See Table 5 in \citet{trella2024oralytics}).

\section{Application Payoff}
\label{sec_app_payoff}
We conduct simulation and re-sampling analyses using data collected during the trial to evaluate design decisions made for our deployed algorithm. We focus on the following questions:

\begin{enumerate}
    \item Was it worth it to invest in fallback methods? (Section~\ref{sec_fallback_impact})
    \item Was it worth it to run a full-pooling algorithm? (Section~\ref{worth_it_to_pool}) 
    \item Despite all these challenges, did the algorithm learn? (Section~\ref{did_we_learn}) 
\end{enumerate}

\subsection{Simulation Environment}
\label{sim_env}
One way to answer questions 2 and 3 is through a simulation environment built using data collected during the Oralytics trial. The purpose of the simulation environment is to re-simulate the trial by generating participant states and outcomes close to the distribution of the data observed in the real trial. This way, we can (1) consider counterfactual decisions (to answer Q2) and (2) have a mechanism for resampling to assess if evidence of learning by the RL algorithm is  due to random chance and thus spurious (to answer Q3). 

For each of the $N=72$ participants with viable data from the trial, we fit a model which is used to simulate OSCB outcomes.
$Q_{i, t}$ given current state $S_{i, t}$ and an action $A_{i, t}$. We also modeled participant app opening behavior and simulated participants starting the trial using the exact date the participant was recruited in the real trial. 
See Appendix~\ref{app_sim_env} 
for full details on the simulation environment.

\begin{figure}[t]  % Specify figure placement: here (h), top (t), bottom (b), page (p)
    \centering
    \includegraphics[width=0.45\textwidth]{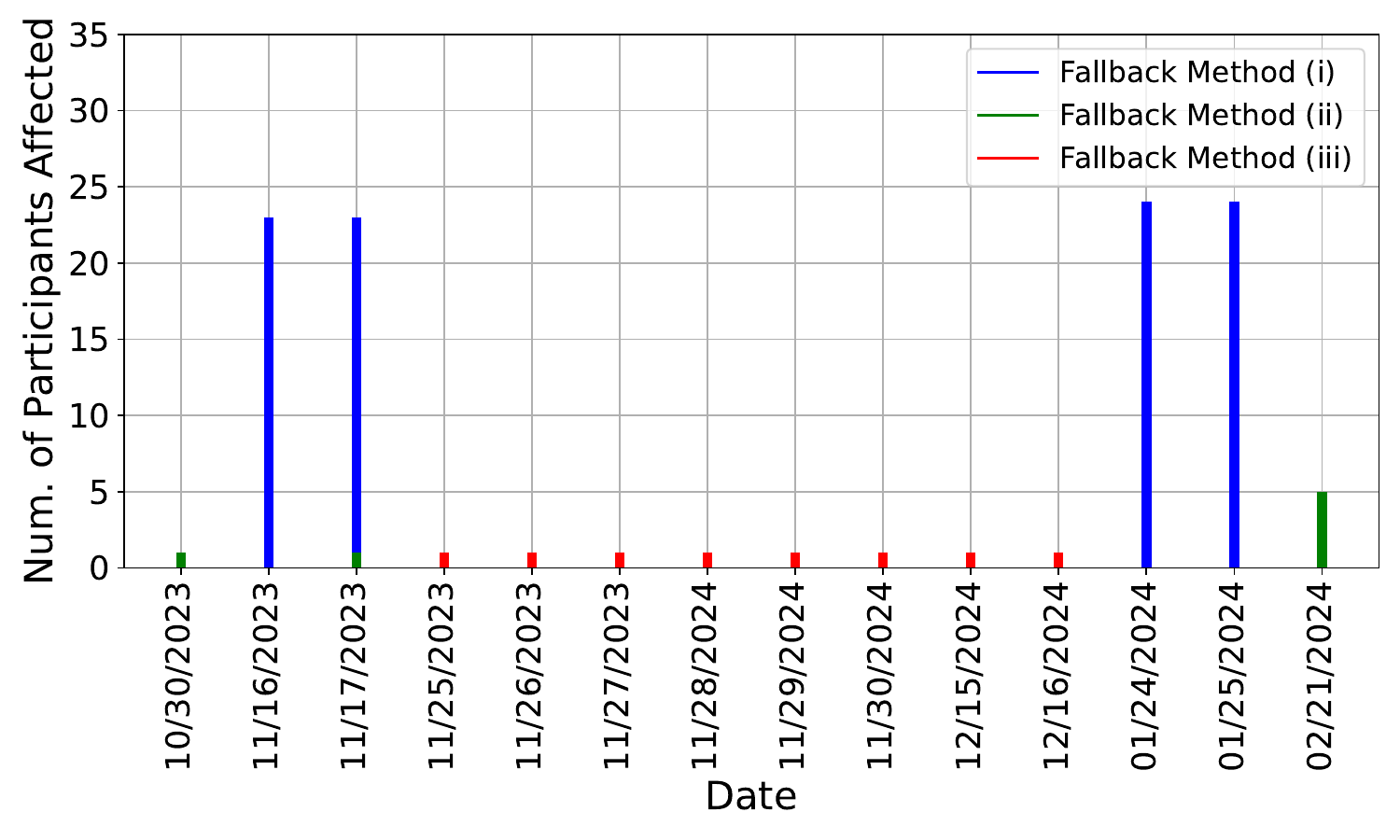}  % Adjust width as needed
    \caption{Fallback methods executed over the Oralytics trial. All 3 fallback methods were executed at least once during the Oralytics trial to mitigate various issues such as the RL service going down or failure to obtain sensor data from the main controller to form current state information.
    }
\label{fig_fb_methods}
\end{figure}

\subsection{Was it worth it to invest in fallback methods?} 
\label{sec_fallback_impact}

\begin{table*}[t]
\begin{tabular}{cllcc}
\hline
Issue ID & Date & Issue Type & Num. Participants Affected & Fallback Method \\
\hline
1 & 10/30/2023 & Fail to read from internal database & 1 & 2 \\
2 & 11/16/2023 & RL Service and endpoints went down & 23 & 1 \\
2 & 11/17/2023 & RL Service and endpoints went down & 23 & 1 \\
3 & 11/17/2023 & Fail to read from internal database & 1 & 2 \\
4 & 11/25/2023 & Fail to get app analytics data from main controller & 1 & 3 \\
4 & 11/26/2023 & Fail to get app analytics data from main controller & 1 & 3 \\
4 & 11/27/2023 & Fail to get app analytics data from main controller & 1 & 3 \\
4 & 11/28/2024 & Fail to get app analytics data from main controller & 1 & 3 \\
4 & 11/29/2024 & Fail to get app analytics data from main controller & 1 & 3 \\
4 & 11/30/2024 & Fail to get app analytics data from main controller & 1 & 3 \\
5 & 12/15/2024 & Fail to get app analytics data from main controller & 1 & 3 \\
5 & 12/16/2024 & Fail to get app analytics data from main controller & 1 & 3 \\
6 & 01/24/2024 & RL Service and endpoints went down & 24 & 1 \\
6 & 01/25/2024 & RL Service and endpoints went down & 24 & 1 \\
7 & 02/21/2024 & Fail to read from internal database & 5 & 2 \\
\hline
\end{tabular}
\caption{Engineering issues that impacted the RL service during the Oralytics trial.}
\label{tab_issues}
\end{table*}
% During the Oralytics trial, all 3 fallback methods were executed at least once.
% One solution to address the transparency and replicability requirements of clinical trials is by implementing fallback methods (Section~\ref{sec_fallback}). 
During the Oralytics trial, various engineering or networking issues (Table~\ref{tab_issues}) occurred that impacted the RL service's intended functionality. These issues were automatically caught and the pre-specified fallback method was executed. Figure~\ref{fig_fb_methods} shows that all 3 types of fallback methods were executed over the Oralytics trial.
Notice that fallback method (i), made possible by our design decision to produce a schedule of actions instead of just a single action, was executed 4 times during the trial and mitigated issues for more participants than any other method.
While defining and implementing fallback methods may take extra effort by the software engineering team, this is a worthwhile investment. Without fallback methods, the various issues that arose during the trial would have required ad hoc changes, to the RL algorithm reducing autonomy and thus replicability of the intervention. 

\subsection{Was it worth it to pool?}
\label{worth_it_to_pool}

\begin{table}[h]
\centering
\begin{tabular}{lccc}
\hline
Pooling &  Mean Value & First Quartile Value \\
\hline
Full Pooling & \textbf{69.724 (0.047)} & \textbf{43.049 (0.091)} \\
No Pooling  & 69.375 (0.047) & 43.024 (0.088) \\
\hline
\end{tabular}
\caption{
Experiment results comparing a full-pooling online RL algorithm with a no-pooling one in the simulation environment.
%Full-pooling achieves higher average and first-quartile OSCB than no-pooling.
Value in each parenthesis is the standard error of the mean across 500 Monte Carlo repetitions.
}
\label{tab_pooling_exps}
\end{table}

Due to the small number of decision points ($T=140$) per participant, the RL algorithm was a full-pooling algorithm (i.e., used a single reward model for all participants and updated using all participants' data). Even though before deployment we anticipated that trial participants would be heterogeneous (i.e., have different outcomes to the intervention), we still believed that full-pooling would learn better over a no-pooling or participant-specific algorithm. Here, we re-evaluate this decision.
% Here, we re-evaluate this decision using the simulation environment that re-simulates the Oralytics trial using participant-specific models for each participant in the trial.

\paragraph{Experiment Setup}
Using the simulation environment (Section~\ref{sim_env}) we re-ran, with all other design decisions fixed as deployed in the Oralytics trial, an algorithm that performs full pooling with one that performs no pooling over $500$ Monte Carlo repetitions. We evaluate algorithms based on:

\begin{itemize}
    \item average of participants' average (across time) OSCB: $$\frac{1}{N}\sum_{i = 1}^N \frac{1}{T}\sum_{t = 1}^T Q_{i, t}$$
    \item first quartile (25th-percentile) of participants' average (across time) OSCB: $$\text{First Quartile}\bigg(\bigg\{\frac{1}{T}\sum_{t = 1}^T Q_{i, t}\bigg\}_{i = 1}^N\bigg)$$
\end{itemize}

\paragraph{Results} As seen in Table~\ref{tab_pooling_exps}, the average and first quartile OSCB achieved by a full-pooling algorithm is slightly higher than the average OSCB achieved by a no-pooling algorithm. These results are congruent with the results for experiments conducted before deployment (Section~\ref{sec_pooling}). Despite the heterogeneity of trial participants, it was worth it to run a full-pooling algorithm instead of a no-pooling algorithm.

\subsection{Did We Learn?}
\label{did_we_learn}

% \begin{figure*}[h!]
%     \centering
%     \subfigure[]{
%     \includegraphics[width=0.45\textwidth]{figs/state_tod_robas+169@developers.pg.com.pdf}
%     }
%     \hfill
%     \subfigure[]{
%         \includegraphics[width=0.45\textwidth]{figs/state_tod_digitaldentalcoach+271@gmail.com.pdf}
%     }
%     \caption{For these two participants, the RL algorithm appears to be personalizing, learning there is an advantage to sending an intervention for state feature time of day = 1 (i.e., evening). We plot action-selection probabilities produced by the RL algorithm for these two participants during the Oralytics trial colored by the time of day state feature (red for morning and blue for evening). 
%     % When the RL algorithm produces action-selection probabilities $> 0.5$, it is for state feature time of day = 1.
%     }
%     \label{fig_tod_personalize}
% \end{figure*}

\begin{table}[t]
\begin{tabular}{c}
\hline
Advantage State Features \\
\hline
    1. Time of Day (Morning/Evening) $\in \{0, 1\}$ \\
    2. Exponential Average of OSCB Over Past Week $\in [-1, 1]$ \\
    3. Exponential Average of Dosage Over Past Week  $\in [-1, 1]$ \\
    4. Prior Day App Engagement $\in \{0, 1\}$
    \\
    5. Intercept Term $=1$ \\
\hline
\end{tabular}
\caption{State features $f(s)$ used by the Oralytics RL algorithm to model the advantage in state $s$. See Appendix~\ref{app_state_space} for more details.
}
\label{tab_state}
\end{table}

% Let the \textit{predicted advantage in state $s$} refer to the algorithm’s prediction of the advantage of action $a = 1$ (i.e., send an engagement prompt) in a particular state $s$. 
% Recall the Oralytics RL algorithm  (Section~\ref{sec_oralytics_rl_alg_overview}),  updates predicted advantages, then via posterior sampling,  selects actions based on the predicted advantages, and  receives feedback/reward. The algorithm uses the received reward to update the  predicted advantage throughout the trial. 
% A natural way to examine what the algorithm learned is to consider the algorithm's prediction or forecast of the benefit of sending an engagement prompt. 

Lastly, we consider if the algorithm was able to learn despite the challenges of the clinical trial setting. We define learning as the RL algorithm successfully learning the advantage of action $a = 1$ over $a = 0$ (i.e., sending an engagement prompt over not sending one) in a particular state $s$. Recall that the Oralytics RL algorithm maintains a model of this advantage (Equation~\ref{eqn_adv_model}) to select actions via posterior sampling and updates the posterior distribution of the advantage model parameters throughout the trial.
One way to determine learning is to visualize the \textit{standardized predicted advantage} in state $s$ throughout the trial (i.e., using learned posterior parameters at different update times $\tau$).
% The standardized predicted advantage in state $s$ is the difference in the posterior mean of rewards in state $s$ for action $a = 1$ versus $a = 0$ standardized using the posterior variance.
The standardized predicted advantage in state $s$ using the policy updated at time $\tau$ is:
\begin{align}
\label{eqn_pred_adv_stat}
    \text{predicted\_adv}(\tau, s) := \frac{\mu^{\beta \top}_{\tau} f(s)}{\sqrt{f(s)^\top \Sigma^{\beta}_\tau f(s)}}
\end{align}
$\mu^{\beta}_{\tau}$ and $\Sigma^{\beta}_{\tau}$ are 
% the sub-vector and sub-matrix of $\mu^{\text{post}}_\tau$ and $\Sigma^{\text{post}}_\tau$ 
the posterior parameters of advantage parameter $\beta$ from Equation~\ref{eqn_adv_model}, and $f(s)$ denotes the features used in the algorithm's model of the advantage (Table~\ref{tab_state}).

\begin{figure}[t]
    \centering
    \includegraphics[width=0.45\textwidth]{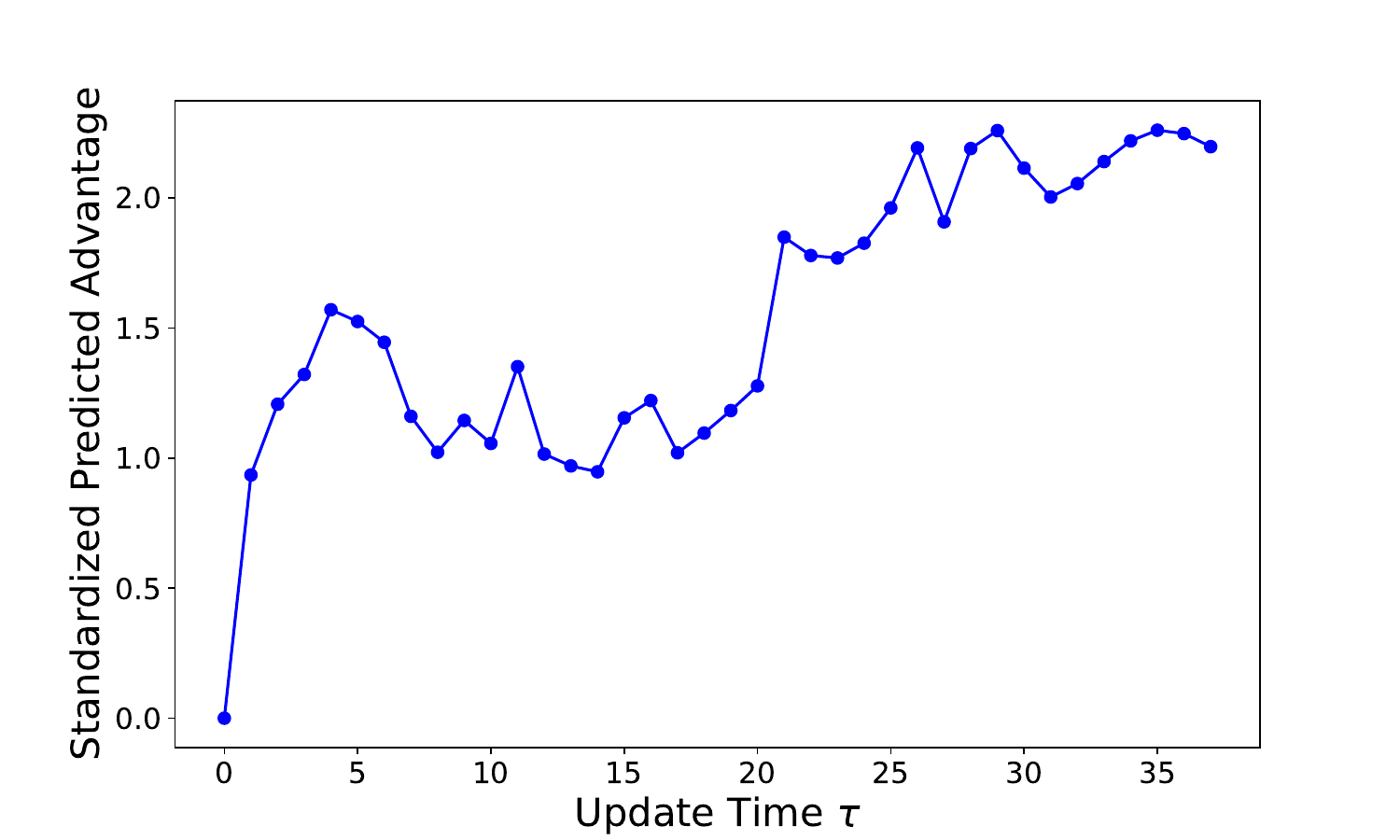}
    \caption{
    The standardized predicted advantage in state $s$ over update times $\tau$ using posterior parameters learned during the Oralytics trial. \textbf{It appears that the algorithm has learned a state where it is effective to send a prompt.}
    }
    \label{fig_appearance_of_learning}
\end{figure}

%%% OLD %%%
% \begin{figure}[t]
%     \centering
%     \includegraphics[width=0.45\textwidth]{figs/posterior_bias.pdf}
%     \caption{
%     % Notice that if the other state features are all 0, then this is the marginal treatment effect according to Susan and Wei Wei
%     The predicted advantage (standardized) in state $s$ over update times $\tau$ using posterior parameters learned during the Oralytics trial. \textbf{It appears that the algorithm has learned it is ineffective to send a prompt in this state.}
%     % This predicted advantage value for $f(s) = [0 , 0, 0, 0, 1]^\top$ corresponds to the algorithm's prediction of the marginal treatment effect.
%     % when participant state is: (1) morning, (2) exponential average OSCB is \sam{near} 90 seconds in the past week (insufficient brushing), (3) participants who received around one prompt a day in the past week (i.e., receive prompts 50\% of the times in the past week), (4) did not open the app the prior day. \sam{suggest to put most of this context in the text as opposed to the legend.  Legend is too long....}\textbf{It appears that the algorithm has learned if the participant is disengaged, then it is ineffective to send an engagement prompt.}
%     }
%     \label{fig_marginal_treatment_eff}
% \end{figure}

For example, consider Figure~\ref{fig_appearance_of_learning}. Using posterior parameters $\mu^{\beta}_{\tau}, \Sigma^{\beta}_{\tau}$ learned during the Oralytics trial, we plot the standardized predicted advantage over updates times $\tau$ in a state where it is
(1) morning, (2) the participant's exponential average OSCB in the past week is about 28 seconds (poor brushing), (3) the participant received prompts 45\% of the times in the past week, and (4) the participant did not open the app the prior day.
Since this value is trending more positive, it {\it appears} that the algorithm learned that it is effective to send an engagement prompt for participants in this particular state. In the following section, we assess whether this  pattern is evidence that the RL algorithm learned or is purely accidental due to the stochasticity in action selection (i.e., posterior sampling).

\begin{figure*}[ht]
    \centering 
    \subfigure[Evening, Poor Brushing, Few Prompts Sent, Not Engaged]{
    % left bottom right top
\includegraphics[width=0.305\textwidth, trim=1cm 0cm 2.5cm 0cm, clip]{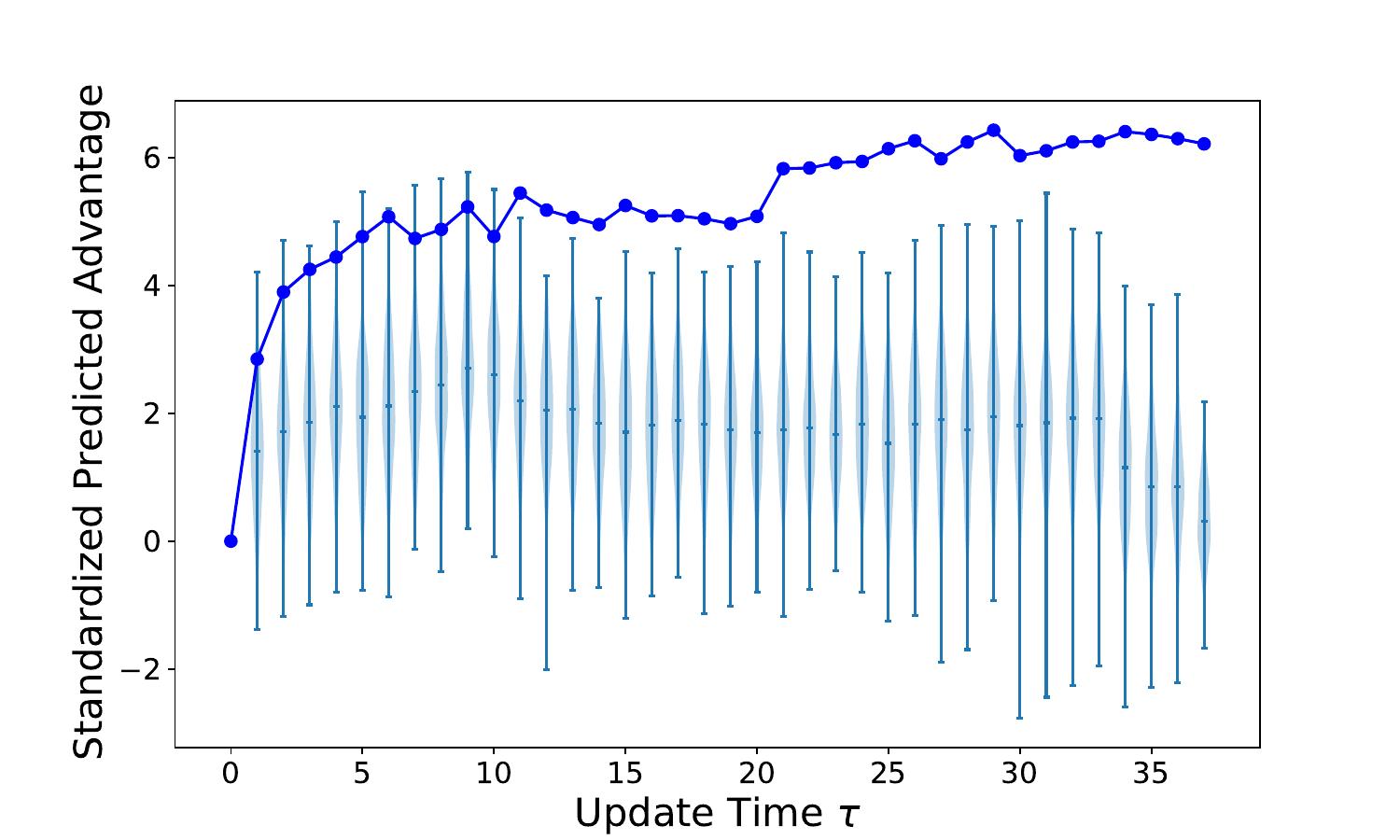} 
    } %[1, -0.7, -0.6, 0, 1]
    \subfigure[Morning, Almost Ideal Brushing, Several Prompts Sent, Engaged]{
        \includegraphics[width=0.32\textwidth, trim=0cm 0cm 2.5cm 0cm, clip]{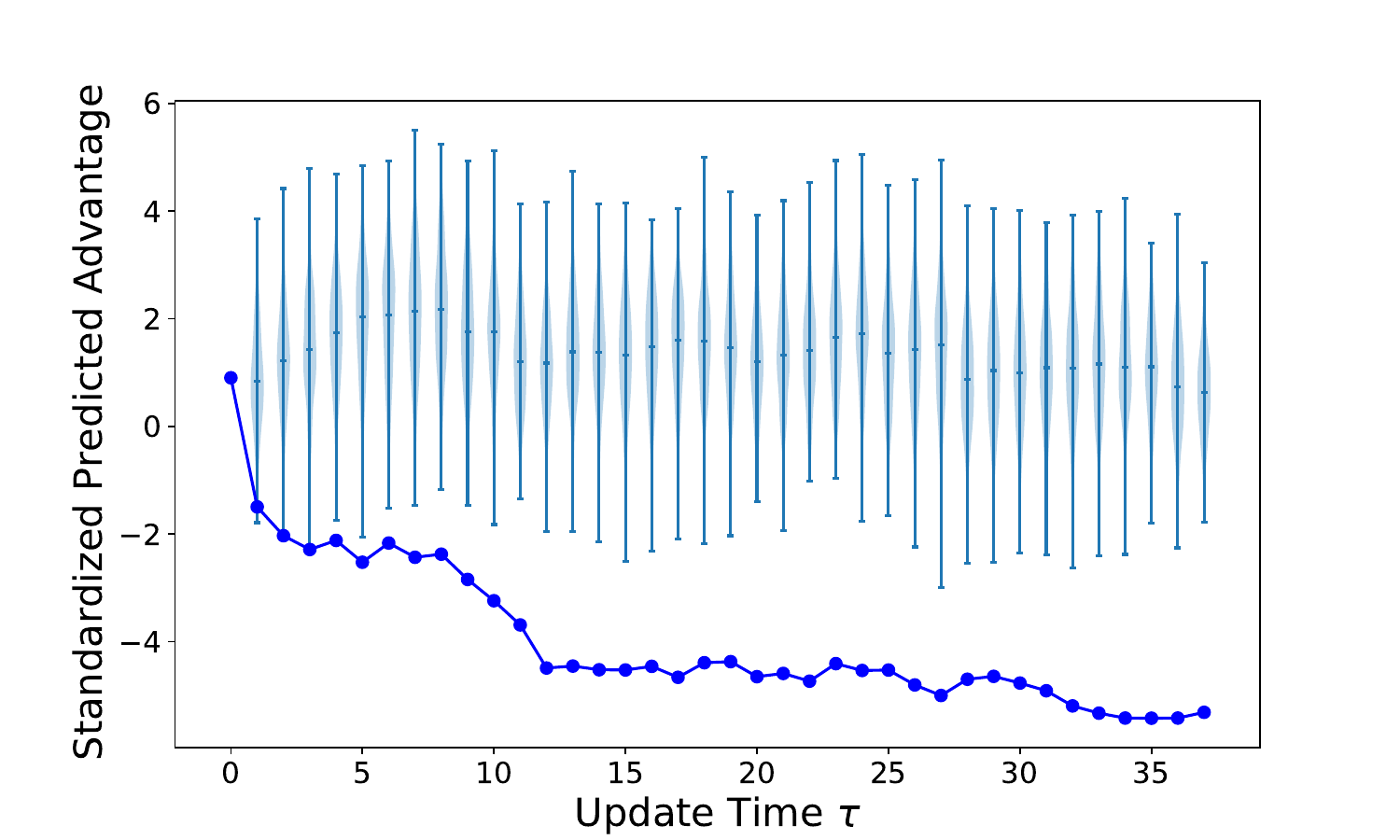} %[0, 0.1, -0.1, 1, 1]
    }
    \subfigure[Morning, Poor Brushing, Several Prompts Sent, Not Engaged]{
        \includegraphics[width=0.32\textwidth, trim=0cm 0cm 2.5cm 0cm, clip]{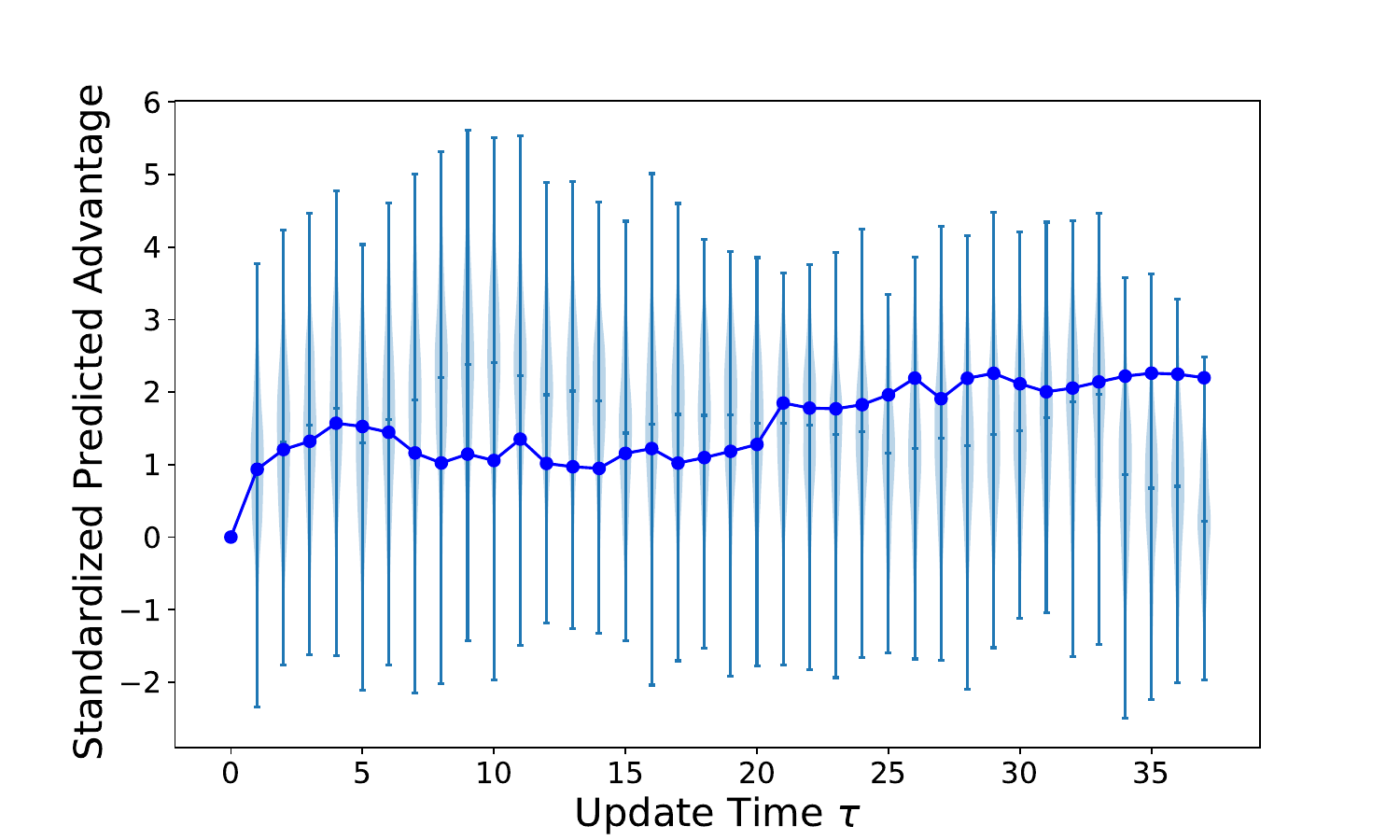} %[1, -0.7, -0.1, 0, 1]
    }
    \caption{
    We compare the standardized predicted advantages across updates to the posterior parameters from the actual Oralytics trial (dark blue) with violin plots of predictive advantages using simulated posterior parameters (light blue) in an environment where there is truly no advantage in state $s$. Simulated posterior parameters were re-sampled across 500 Monte Carlo repetitions. The pattern in (a) and (b) suggests states where the algorithm learned an advantage of one action over the other and the re-sampling indicates this evidence is real. The pattern in (c), however, suggests a state where re-sampling indicates the appearance of learning likely occurred by random chance.
    }
    % Standardized predicted advantages across update times $\tau$ across 500 Monte Carlo repetitions. Each repetition represents a simulated trial composed of $N=72$ resampled participants that generate OSCB where there is no advantage of action $1$ in state $s$. The actual predictive advantage from the Oralytics trial is in dark blue.
    \label{fig_resampling_results}
\end{figure*}

\paragraph{Experiment Setup}
We use the re-sampling-based parametric method developed in \citet{ghosh2024did} to assess if the evidence of learning could have occurred by random chance. We use the simulation environment built using the Oralytics trial data (Section~\ref{sim_env}).
For each state of interest $s$, we run the following simulation.
(i) We rerun the RL algorithm in a variant of the simulation environment in which there is \textit{no advantage} of action $1$ over action $0$ in state $s$ (See Appendix~\ref{app_resampling_method}) producing posterior means and variances, $\mu^{\beta}_{\tau}$ and $\Sigma^{\beta}_{\tau}$. 
Using $\mu^{\beta}_{\tau}$ and $\Sigma^{\beta}_{\tau}$, we calculate standardized predicted advantages for each update time $\tau$. (ii) We compare the standardized predicted advantage (Equation~\ref{eqn_pred_adv_stat}) at each update time  from the real trial 
with the standardized predicted advantage from the simulated trials in (i).

We consider a total of 16 different states of interest. To create these 16 states, we consider different combinations of possible values for algorithm advantage features $f(s)$ (Table~\ref{tab_state}). Features (1) and (4) are binary so we consider both values $\{0, 1\}$ for each. Features (2) and (3) are real-valued between $[-1, 1]$, so we consider the first and third quartiles calculated from the Oralytics trial data.\footnote{For feature (2), -0.7 corresponds to an exponential average OSCB in the past week of 28 seconds and 0.1 corresponds to 100 seconds; for feature (3), -0.6 corresponds to the participant receiving prompts 20\% of the time in the past week and -0.1 corresponds to 45\%.} 

\paragraph{Results}
Key results are in Figure~\ref{fig_resampling_results} and additional plots are in Appendix~\ref{app_additional_figs}.
Our results show that the Oralytics RL algorithm did indeed learn that sending a prompt is effective in some states and ineffective in others. This suggests that our state space design was a good choice because some state features helped the algorithm discern these states.

% We highlight 3 interesting states in Figure~\ref{fig_resampling_results}: (a) a state where the algorithm learned it is effective to send a prompt and the re-sampling indicates this evidence is real, (b) a state where the algorithm learned it is ineffective to send a prompt and the re-sampling indicates this evidence is real, and (c) the state in Figure~\ref{fig_appearance_of_learning} but the re-sampling method indicates the appearance of learning likely occurred by random chance. 
We highlight 3 interesting states in Figure~\ref{fig_resampling_results}: 
\begin{enumerate}[label=(\alph*)]
    \item A state where the algorithm learned it is effective to send a prompt and the re-sampling indicates this evidence is real. The advantage features $f(s)$ correspond to (1) evening, (2) the participant’s exponential average OSCB in the past week is about 28 seconds (poor brushing),
(3) the participant received prompts 20\% of the time in the past week, and (4) the participant did not open the app the prior day.
    \item A state where the algorithm learned it is ineffective to send a prompt and the re-sampling indicates this evidence is real. The advantage features $f(s)$ correspond to (1) morning, (2) the participant’s exponential average OSCB in the past week is about 100 seconds (almost ideal brushing),
(3) the participant received prompts 45\% of the time in the past week, and (4) the participant opened the app the prior day.
    \item The state in Figure~\ref{fig_appearance_of_learning} but the re-sampling method indicates the appearance of learning likely occurred by random chance. 
\end{enumerate}

For (a) and (b) the re-sampling method suggests that evidence of learning is real because predicted advantages using posterior parameters updated during the actual trial are trending away from the simulated predictive advantages from re-sampled posterior parameters in an environment where there truly is no advantage in state $s$. For (c), however, the re-sampling method suggests that the appearance of learning likely occurred by random chance because predicted advantages using posterior parameters updated during the actual trial are extremely similar to those from re-sampled posterior parameters in an environment where there truly is no advantage in state $s$.

\section{Discussion}
We have deployed Oralytics, an online RL algorithm optimizing prompts to improve oral self-care behaviors.
As illustrated here, much is learned from the end-to-end development, deployment, and data analysis phases.
We share these insights by highlighting design decisions for the algorithm and software service and conducting a simulation and re-sampling analysis to re-evaluate these design decisions using data collected during the trial. Most interestingly, the re-sampling analysis provides evidence that the RL algorithm learned the advantage of one action over the other in certain states. 
We hope these key lessons can be shared with other research teams interested in real-world design and deployment of online RL algorithms.
From a health science perspective, pre-specified, primary analyses \cite{nahum2024optimizing} will occur, which is out of scope for this paper. The re-sampling  analyses presented in this paper will inform design decisions for phase 2. The re-design of the RL algorithm for phase 2 of the Oralytics clinical trial  is currently under development and phase 2 is anticipated to start in spring 2025. 
% We are currently re-designing the RL algorithm for the next deployment in phase 2 of the Oralytics clinical trial anticipated for spring 2025.

% Please try to limit acknowledgments to no more than three sentences.

\section*{Acknowledgments}
This research was funded by NIH grants IUG3DE028723, P50DA054039, P41EB028242, U01CA229437, UH3DE028723, and R01MH123804. SAM  holds concurrent appointments  at Harvard University and as an Amazon Scholar. This paper describes work performed at Harvard University and is not associated with Amazon.

% \section*{Ethical Statement.}
% You can write a statement about the potential ethical impact of your work, including its broad societal implications, both positive and negative. If included, such statement must be written in an unnumbered section titled \emph{Ethical Statement}.

%\clearpage
\bibliography{main}

\appendix
\onecolumn
% \section{Simulation Results}

% % \begin{table}[h]
% % \centering
% % \caption{Simulation Results}
% \label{tab:evaluation_results}
% \begin{tabular}{lccc}
% \hline
% Pooling & Online vs. Offline & Mean Value & Low 25th Percentile \\
% \hline
% Full Pooling & Offline & 17.625 (0.013) & 8.352 (0.038) \\
% Full Pooling & Online  & 17.530 (0.014) & 8.342 (0.037) \\
% Full Pooling & Online  & 17.530 (0.014) & 8.342 (0.037) \\
% No Pooling  & Online  & 17.691 (0.014) & 8.415 (0.036) \\
% \hline
% \end{tabular}
% \end{table}

\section{Additional Oralytics RL Algorithm Facts}
\subsection{Algorithm State Space}
\label{app_state_space}
$S_{i,t} \in \mathbb{R}^d$ represents the $i$th participant's state at decision point $t$, where $d$ is the number of variables describing the participant's state. 

\subsubsection{Baseline and Advantage State Features}
% $f(S_{i,t}) \in \mathbb{R}^5$ denotes the features used to model both the baseline reward function and the advantage. Notice that for Oralytics, the baseline and advantage features are the same (i.e., all $f(S_{i,t})$), but this is a design choice, and they do not have to be.
Let $f(S_{i,t}) \in \mathbb{R}^5$ denote the features used in the algorithm's model for both the baseline reward function and the advantage. 
% In the case of Oralytics, the features defined in $f(S_{i, t})$ are the same features used to model both the baseline reward and the advantage.

\paragraph{} These features are:
\begin{enumerate}
\label{alg_state_features}
    \item Time of Day (Morning/Evening) $\in \{0, 1\}$
    \item \label{alg_state:brushing} $\Bar{B}$: Exponential Average of OSCB Over Past 7 Days (Normalized) $\in [-1, 1]$
    \item \label{alg_state:actions} $\Bar{A}$: Exponential Average of Engagement Prompts Sent Over Past 7 Days (Normalized)  $\in [-1, 1]$
    \item \label{alg_state:app} Prior Day App Engagement $\in \{0, 1\}$
    \item Intercept Term $=1$
\end{enumerate}
Feature 1 is 0 for morning and 1 for evening. Features \ref{alg_state:brushing} and \ref{alg_state:actions} are $\bar{B}_{i,t} = c_{\gamma}\sum_{j = 1}^{14} \gamma^{j-1} Q_{i, t - j}$ and $\bar{A}_{i,t} = c_{\gamma}\sum_{j = 1}^{14} \gamma^{j-1} A_{i, t - j}$ respectively, where $\gamma=13/14$ and $c_{\gamma} = \frac{1 - \gamma}{1 - \gamma^{14}}$. Recall that $Q_{i, t}$ is the proximal outcome of OSCB and $A_{i,t}$ is the treatment indicator. Feature 4 is 1 if the participant has opened the app in focus (i.e., not in the background) the prior day and 0 otherwise. Feature 5 is always 1. For full details on the design of the state space, see Section 2.7 in \citet{trella2024oralytics}.

\subsection{Reward Model}
\label{app_reward_model}
The reward model (i.e., model of the mean reward given state $s$ and action $a$) used in the Oralytics trial is a Bayesian linear regression model with action centering \cite{DBLP:journals/corr/abs-1909-03539}:

\begin{equation}
\label{eqn:blr}
    r_{\theta}(s, a) = f(s)^T \alpha_0 + \pi f(s)^T \alpha_1 + (a - \pi) f(s)^T \beta + \epsilon
\end{equation}
where $\theta = [\alpha_0, \alpha_1, \beta]$ are model parameters, $\pi$ is the probability that the RL algorithm selects action $a = 1$ in state $s$
% $S_{i,t}$ for participant $i$ at decision point $t$
and $\epsilon \sim \mathcal{N}(0, \sigma^2)$. We call the term $f(S_{i, t})^T \beta$ the advantage (i.e., advantage of selecting action 1 over action 0) and $f(S_{i, t})^T \alpha_0 + \pi_{i,t} f(S_{i, t})^T \alpha_1$ the baseline.
The priors are $\alpha_{0} \sim \mathcal{N}(\mu_{\alpha_0}, \Sigma_{\alpha_0})$, $\alpha_{1} \sim \mathcal{N}(\mu_{\beta}, \Sigma_{\beta})$, $\beta \sim \mathcal{N}(\mu_{\beta}, \Sigma_{\beta})$. Prior values for $\mu_{\alpha_0}, \Sigma_{\alpha_0}, \mu_{\beta}, \Sigma_{\beta}, \sigma^2$ are specified in Section~\ref{app_prior}. For full details on the design of the reward model, see Section 2.6 in \citet{trella2024oralytics}.

\subsection{Prior}
\label{app_prior}
% \sam{provide prior distribution and reference \citet{trella2024oralytics} for how the prior was constructed.}
Table~\ref{tab:finalized_prior} shows the prior distribution values used by the RL algorithm in the Oralytics trial. For full details on how the prior was constructed, see Section 2.8 in \citet{trella2024oralytics}.

\begin{table*}[!ht]
    \centering
    \begin{tabular}{c|c}
    \toprule
        Parameter & Oralytics Pilot \\
        \hline
        \midrule
        $\sigma^2$: noise variance & 3878 \\
        $\mu_{\alpha_0}$: prior mean of the baseline state features & $[18, 0, 30, 0, 73]^T$ \\
        $\Sigma_{\alpha_0}$: prior variance of the baseline state features & $\text{diag}(73^2, 25^2, 95^2, 27^2, 83^2)$ \\
        $\mu_{\beta}$: prior mean of the advantage state features & $[0, 0, 0, 53, 0]^T$ \\
        $\Sigma_{\beta}$: prior variance of the advantage state features & $\text{diag}(12^2, 33^2, 35^2, 56^2, 17^2)$ \\
    \end{tabular}
    \caption{\textbf{Prior Used in Oralytics Trial.} Values are rounded to the nearest integer. Recall that the ordering of the features is the same as described in Section~\ref{alg_state_features}: Time of Day, Exponential Average of Brushing Over Past 7 Days (Normalized), Exponential Average of Engagement Prompts Sent Over Past 7 Days (Normalized), Prior Day App Engagement, Intercept Term.}
    \label{tab:finalized_prior}
\end{table*}

\section{Simulation Environment}
\label{app_sim_env}
We created a simulation environment using the Oralytics trial data in order to replicate the trial under different true environments. 
Although the trial ran with 79 participants, due to an engineering issue, data for 7 out of the 79
participants was incorrectly saved and thus their data is unviable. Therefore, the simulation environment is built off of data from the 72 unaffected participants.
Replications of the trial are useful to (1) re-evaluate design decisions that were made and (2) have a mechanism for resampling to assess if evidence of learning by the RL algorithm is due to random chance. 
For each of the 72 participants with viable data from the Oralytics clinical trial, we use that participant's data to create a participant-environment model. We then re-simulate the Oralytics trial by generating participant states, the RL algorithm selecting actions for these 72 participants given their states, the participant-environment model generating health outcomes / rewards in response, and the RL algorithm updating using state, action, and reward data generated during simulation. To make the environment more realistic, we also replicate each participant being recruited incrementally and entering the trial by their real start date in the Oralytics trial and simulate update times on the same dates as when the RL algorithm updated in the real trial (i.e., weekly on Sundays). 

\subsection{Participant-Environment Model}
In this section, we describe how we constructed the participant-environment models for each of the $N = 72$ participants in the Oralytics trial using that participant's data. Each participant-environment model has the following components:
\begin{itemize}
    \item Outcome Generating Function (i.e., OSCB $Q_{i, t}$ in seconds given state $S_{i, t}$ and action $A_{i, t}$)
    \item App Engagement Behavior (i.e., the probability of the participant opening their app on any given day)
\end{itemize}

\paragraph{Environment State Features} The %baseline and advantage 
features used in the state space for each environment are 
% time-varying features are 
a superset of the algorithm state features $f(S_{i, t})$ (Appendix~\ref{app_state_space}). $g(S_{i,t}) \in \mathbb{R}^7$ denotes the super-set of features used in the environment model. %for both the baseline outcome and the advantage.

The features are:
\begin{enumerate}
\label{env_state_features}
    \item Time of Day (Morning/Evening) $\in \{0, 1\}$
    \item \label{alg_state:brushing2} $\Bar{B}$: Exponential Average of OSCB Over Past 7 Days (Normalized) $\in [-1, 1]$
    \item \label{alg_state:actions2} $\Bar{A}$: Exponential Average of Prompts Sent Over Past 7 Days (Normalized)  $\in [-1, 1]$
    \item \label{alg_state:app2} Prior Day App Engagement $\in \{0, 1\}$
    \item Day of Week (Weekend / Weekday) $\in \{0, 1\}$
    \item Days Since Participant Started the Trial (Normalized) $\in [-1, 1]$
    \item Intercept Term $=1$
\end{enumerate}
%Features 5 and 6 are features we use in the participant-environment model, but not features used by the RL algorithm. 
Feature 5 is 0 for weekdays and 1 for weekends. Feature 6 refers to how many days the participant has been in the Oralytics trial (i.e., between 1 and 70) normalized to be between -1 and 1.

\paragraph{Outcome Generating Function}
The outcome generating function is a function that generates OSCB $Q_{i, t}$ in seconds given current state $S_{i, t}$ and action $A_{i, t}$. We use a zero-inflated Poisson to model each participant's outcome generating process because of the zero-inflated nature of OSCB found in previous data sets and data collected in the Oralytics trial. Each participant's outcome generating function is:

\begin{align*}
    Z \sim \text{Bernoulli} \bigg(1 - \mathrm{sigmoid} \big( g(S_{i, t})^\top w_{i,b} - A_{i, t} \cdot \max \big[ \Delta_{i,B}^\top g(S_{i, t}), 0 \big] \big) \bigg)
\end{align*}
\begin{align}
\label{eqn_outcome_generating_func}
    S \sim \text{Poisson} \big( \exp \big( g(S_{i, t})^\top w_{i,p} + A_{i, t} \cdot \max\big[ \Delta_{i,N}^\top g(S_{i, t}), 0 \big] \big) \big)
\end{align}
\begin{align*}
    Q_{i, t} = ZS
\end{align*}
where $g(S_{i, t})^\top w_{i,b},g(S_{i, t})^\top w_{i,p}$ are called baseline (aka when $A_{i,t}=0$) models with 
$w_{i,b}, w_{i,p}$ as participant-specific baseline weight vectors, $\max \big[ \Delta_{i,B}^\top g(S_{i, t}), 0 \big], \max\big[ \Delta_{i,N}^\top g(S_{i, t}), 0 \big]$ are called advantage models, with $\Delta_{i,B}, \Delta_{i,N}$ as participant-specific advantage (or treatment effect) weight vectors. $g(S_{i, t})$ is  described in Appendix~\ref{env_state_features}, and $\mathrm{sigmoid}(x) = \frac{1}{1 + e^{-x}}$.

% interpretability of model
The outcome generating function can be interpreted in two components: (1) the Bernoulli outcome $Z$ models the participant's intent to brush given state $S_{i, t}$ and action $A_{i, t}$ and (2) the Poisson outcome $S$ models the participant's OSCB value in seconds when they intend to brush, given state $S_{i, t}$ and action $A_{i, t}$. Notice that the models for $Z$ and $S$ currently require the advantage/treatment effect of OSCB $Q_{i, t}$ to be non-negative. Otherwise, sending an engagement prompt would yield a lower OSCB value (i.e., models participant brushing worse) than not sending one, which was deemed nonsensical in this mHealth setting.
%We ensure that $\max\big[ \Delta_{i,B}^\top h(S_{i, t}), 0 \big]$ and $\max\big[ \Delta_{i,N}^\top h(S_{i, t}), 0 \big]$ are non-negative to prevent the treatment effect  from switching signs and having a negative effect on OSCB.

Weights $w_{i,b}, w_{i,p}, \Delta_{i,B}, \Delta_{i,N}$ for each participant's outcome generating function are fit that participant's state, action, and OSCB data from the Oralytics trial. We fit the function using MAP with priors $w_{i,b}, w_{i,p}, \Delta_{i,B}, \Delta_{i,N} \sim \mathcal{N}(0, I)$ as a form of regularization because we have sparse data for each participant. Finalized weight values were chosen by running random restarts and selecting the weights with the highest log posterior density. See Appendix~\ref{app_assess_env_quality} for metrics calculated to verify the quality of each participant's outcome generating function.

\paragraph{App Engagement Behavior}
We simulate participant app engagement behavior using that participant's app opening data from the Oralytics trial. Recall that app engagement behavior is used in the state for both the environment and the algorithm. More specifically, we define app engagement as the participant opening their app and the app is in focus and not in the background. Using this app opening data, we calculate $p^{\text{app}}_i$, the proportion of days that the participant opened the app during the Oralytics trial (i.e., number of days the participant opened the app in focus divided by $70$, the total number of days a participant is in the trial for). During simulation, at the end of each day, we sample from a Bernoulli distribution with probability $p^{\text{app}}_i$ for every participant $i$ currently in the simulated trial.

\subsection{Assessing the Quality of the Outcome Generating Functions}
\label{app_assess_env_quality}
\begin{figure}[!h]
    \centering
        \centering
\includegraphics[width=\textwidth]{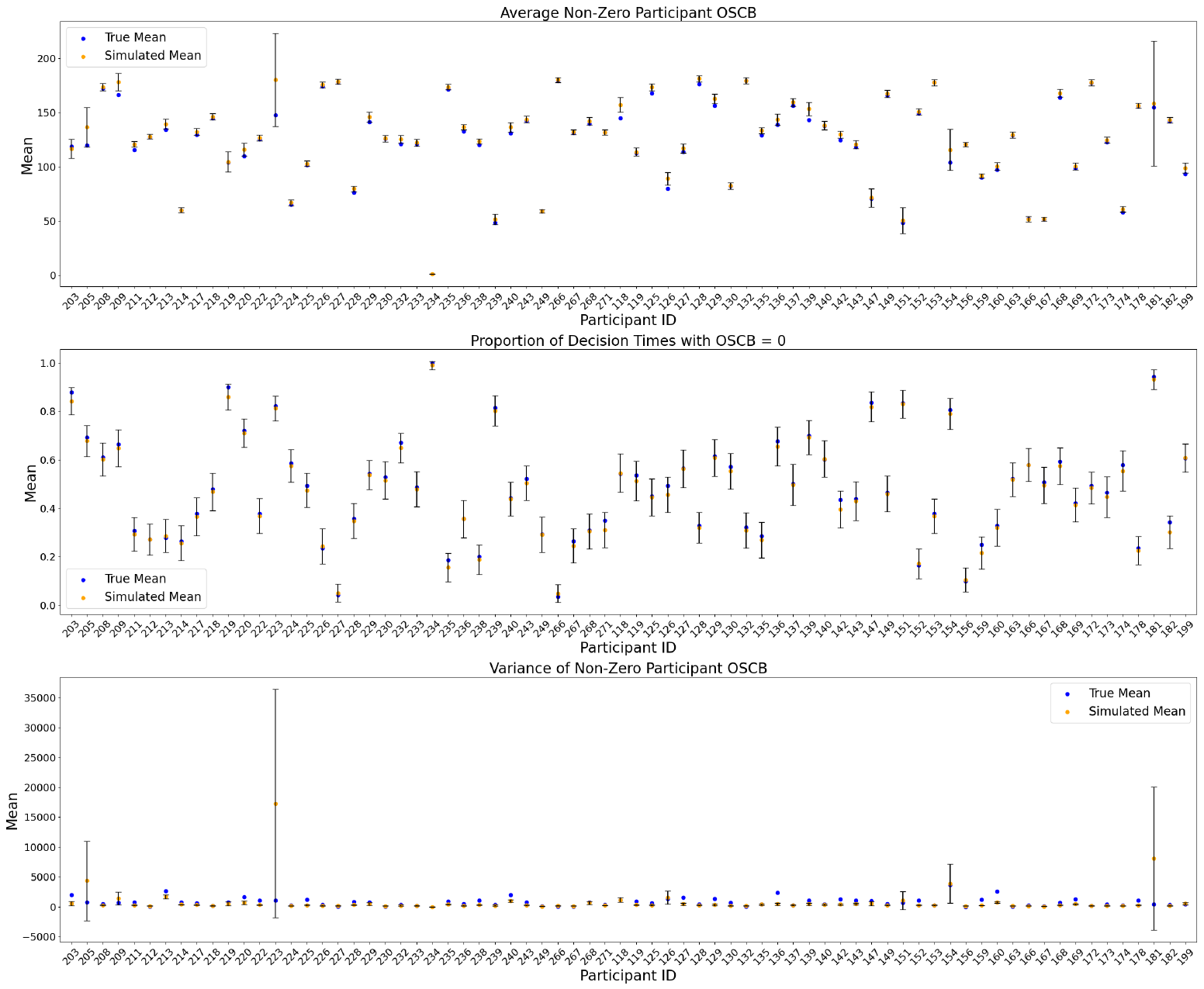}
    \caption{Outcome metrics across trial participants comparing data generated by the simulation environment with Oralytics trial data. Error bars depict confidence intervals across 500 Monte Carlo repetitions.
    }
    \label{fig_oralytics_app_ex}
\end{figure}

Our goal is to have the simulation environment replicate outcomes (i.e., OSCB) as close to the real Oralytics trial data as possible. To verify this, we compute various metrics (defined in the following section) comparing how close the outcome data generated by the simulation environment is to the data observed in the real trial . Table~\ref{tab:sim_vs_oralytics} shows this comparison on various outcome metrics.
Table~\ref{tab_sim_env_err_vals} shows various error values of simulated OSCB with OSCB observed in the trial.
For both tables, we report the average and standard errors of the metric across the 500 Monte Carlo simulations and compare with the value of the metric for the Oralytics trial data. Figure~\ref{fig_oralytics_app_ex} shows comparisons of outcome metrics across trial participants.

\subsubsection{Notation} 
$\mathbb{I}\{\cdot\}$ denotes the indicator function. Let $\widehat{\text{Var}}(\{X_k\}_{k = 1}^K)$ represent the empirical variance of $X_1,...,X_K$.

\subsubsection{Metric Definitions and Formulas}
Recall that $N=72$ is the number of participants and $T=140$ is the total number of decision times that the participant produces data for in the trial. We consider the following metrics and compare the metric on the real data with data generated by the simulation environment.

\begin{enumerate}
    \item Proportion of Decision Times with OCSB = 0:
\begin{align}
\label{metric_1}
\frac{\sum_{i=1}^{N} \sum_{t=1}^{T}\mathbb{I}\{Q_{i,t} = 0\}}{N \times T}
\end{align}

\item Average of Average Non-zero Participant OSCB:
% average of average Nonzero OSCB
\begin{align}
\label{metric_2}
\frac{1}{N} \sum_{i=1}^{N} \bar{Q}_i^{\text{non-zero}}
\end{align}

where $$\bar{Q}_i^{\text{non-zero}} = \frac{\sum_{t=1}^{T} Q_{i,t} \cdot \mathbb{I}\{Q_{i,t} > 0\}}{\sum_{t=1}^{T} \mathbb{I}\{Q_{i,t} > 0\}}$$

\item Average Non-zero OSCB in Trial:
% pooling all participant's data together to form one giant data set and average over non-zero OSCB in that giant data set
\begin{align}
\label{metric_3}
    \frac{1}{\sum_{i = 1}^N \sum_{t = 1}^T \mathbb{I}\{Q_{i, t} > 0\}}  \sum_{i = 1}^N \sum_{t = 1}^T Q_{i, t} \cdot \mathbb{I}\{Q_{i, t} > 0\}
\end{align}

\item Variance of Average Non-zero Participant OSCB:
\begin{align}
\label{metric_4}
\widehat{\text{Var}}(\{\bar{Q}_i^{\text{non-zero}}\}_{i = 1}^N)
\end{align}

where $$\bar{Q}_i^{\text{non-zero}} = \frac{\sum_{t=1}^{T} Q_{i,t} \cdot \mathbb{I}\{Q_{i,t} > 0\}}{\sum_{t=1}^{T} \mathbb{I}\{Q_{i,t} > 0\}}$$

% where 
% \begin{align*}
% \text{Var}(Q_k^{\text{non-zero}}) = \frac{\sum_{t=1}^{T} \left(Q_{k,t} - \bar{Q}_k^{\text{non-zero}}\right)^2 \cdot \mathbb{I}\{Q_{k,t} > 0\}}{\sum_{i=1}^{N} \sum_{t=1}^{T} \mathbb{I}\{Q_{i,t} > 0\}}
% \end{align*}

\item Variance of Non-zero OSCB in Trial: 
% pooling all participant's data together to form one giant data set and taking the variance over non-zero OSCB in that giant data set
\begin{align}
\label{metric_5}
    \widehat{\text{Var}}(\{Q_{i, t} : Q_{i, t} > 0\}_{i = 1, t = 1}^{N, T})
\end{align}
% \begin{align*}
% \frac{\sum_{i=1}^{N} \sum_{t=1}^{T} \left(Q_{i,t} - \mu\right)^2 \mathbb{I}\{Q_{i,t} > 0\}}{\sum_{i=1}^{N} \sum_{t=1}^{T} \mathbb{I}\{Q_{i,t} > 0\}}
% \end{align*}

\item Variance of Average Participant OCSB:
%  This metric measures the degree of between-participant variance in average brushing
\begin{align}
\label{metric_6}
 \widehat{\text{Var}}(\{\bar{Q}_i\}_{i = 1}^N)
\end{align}

where $\bar{Q}_i = \sum_{t = 1}^T Q_{i, t}$ is the average OSCB for participant $i$
\\

\item Average of Variances of Participant OSCB:
%  This metric measures the average amount of within-participant variance.
\begin{align}
\label{metric_7}
\frac{1}{N} \sum_{i=1}^{N} \widehat{\text{Var}}(\{Q_{i, t}\}_{t = 1}^T) 
\end{align}

% where $\text{Var}_k$ is the variance of quality within each participant, ie. $$\text{Var}(Q_k) = \frac{1}{N\times T} \sum_{i=1}^{N \times T} (Q_{k,i} - \bar{Q}_k)^2$$
% \\
\end{enumerate}

We also compute the following error metrics. We use $\hat{Q}_{i, t}$ to denote the simulated OSCB and $Q_{i, t}$ to denote the corresponding OSCB value from the Oralytics trial data. 

\begin{enumerate}
\item Mean Squared Error:
\begin{align}
\label{eqn_mse}
    \frac{1}{N\times T} \sum_{i=1}^{N} \sum_{t=1}^{T}(\hat{Q}_{i,t} - Q_{i,t})^2
\end{align}
\item Root Mean Squared Error: 
\begin{align}
\label{eqn_rmse}
    \sqrt{\frac{1}{N\times T} \sum_{i=1}^{N} \sum_{t=1}^{T}(\hat{Q}_{i,t} - Q_{i,t})^2}
\end{align}
\item Mean Absolute Error: 
\begin{align}
\label{eqn_mean_abs_err}
    \frac{1}{N\times T} \sum_{i=1}^{N} \sum_{t=1}^{T}|\hat{Q}_{i,t} - Q_{i,t}|
\end{align}
\end{enumerate}

\begin{table}[!h]
\centering
\begin{tabular}{lrr}
\hline
\hline
Outcome Metric & Simulation Environment & Oralytics Trial Data \\
\hline
\hline
Proportion of Decision Times With OSCB = 0 (Equation~\ref{metric_1}) & 0.473 (0.0002) & 0.477 \\
Average Non-Zero OSCB in Trial (Equation~\ref{metric_2}) & 131.196 (0.018) & 131.487 \\
Average of Average Non-Zero Participant OSCB (Equation~\ref{metric_3}) & 126.894 (0.043) & 127.104 \\
Variance of Non-Zero OSCB in Trial (Equation~\ref{metric_4}) & 1790.723 (3.208) & 1777.210 \\
Variance of Average Non-Zero Participant OSCB (Equation~\ref{metric_5}) & 834.028 (7.434) & 796.132 \\
Variance of Average Participant OSCB (Equation~\ref{metric_6})& 69.166 (0.024) & 68.827 \\
Average of Variances of Participant OSCB (Equation~\ref{metric_7}) & 3865.696 (2.723) & 3883.210 \\
\hline
\end{tabular}
\caption{Outcome metrics for data from generated from the simulation environment vs. Oralytics trial data.
Each outcome metric value under the ``Simulation Environment" column is computed for each of the 500 Monte Carlo simulated repetitions. We report the mean (rounded to nearest 3 decimal places) and the standard errors (in parentheses) of these metrics across the repetitions.
% \hsj{Anna, I have confirmed that the simulation env. values are averages of 500 repetitions and reported standard errors. Also, the metric names match the names used above.}
}
\label{tab:sim_vs_oralytics}
\end{table}

\begin{table}[h]
\centering
\begin{tabular}{ll}
\hline
\hline
Error Metric & Value \\
\hline
\hline
Mean Squared Error (Equation~\ref{eqn_mse})& 6165.169 (4.485) \\
Root Mean Squared Error (Equation~\ref{eqn_rmse}) & 78.516 (0.029)\\
Mean Absolute Error (Equation~\ref{eqn_mean_abs_err})& 48.027 (0.023)\\
\bottomrule
\end{tabular}
\caption{Simulation Environment Error Values. Error values are computed using the simulated OSCB and the OSCB values in the Oralytics trial data. An error value is computed for each of the 500 Monte Carlo repetitions. We report the mean (rounded to nearest 3 decimal places) and the standard errors (in parentheses) across these repetitions.
}
\label{tab_sim_env_err_vals}
\end{table}

% \begin{itemize}
%     \item Comparing the average brushing quality in the real trial vs. simulated brushing quality
%     \item Plot a histogram of all brushing qualities in the real trial vs. something that we need to think about.
%     \item Calculate a between-user variance (of outcomes) between $72$ participants and calculate
%     \item Calculate a within-user variance (of outcomes) between $140$ decision times for each of the $72$ participants
%     \item Check realistic effect sizes: compare the standardized effect size that we see in the real trial vs. simulated trial
% \end{itemize}

\newpage
\subsection{Environment Variants for Re-sampling Method}
\label{app_resampling_method}
In this section, we discuss how we formed variants of the simulation environment used in the re-sampling method from Section~\ref{did_we_learn}. We create a variant for every state $s$ of interest corresponding to algorithm advantage features $f(s)$ and environment advantage features $g(s)$. In each variant, outcomes (i.e., OSCB $Q_{i, t}$) and therefore rewards, are generated so that there is \textit{no advantage} of action $1$ over action $0$ in the particular state $s$. 
% Notice this means there could be an advantage in other states $s'$.

To do this, recall that we fit an outcome generating function (Equation~\ref{eqn_outcome_generating_func}) for each of the $N = 72$ participants in the trial. Each participant $i$'s outcome generating function has advantage weight vectors $\Delta_{i,B}, \Delta_{i,N}$ that interact with the environment advantage state features $g(s)$. Instead of using $\Delta_{i,B}, \Delta_{i,N}$ fit using that participant's trial data, we instead use projections $\text{proj} \; \Delta_{i,B}, \text{proj} \; \Delta_{i,N}$ of $\Delta_{i,B}, \Delta_{i,N}$ that have two key properties:
\begin{enumerate}
    \item for the current state of interest $s$, on average they generate treatment effect values that are 0 in state $s$ with algorithm state features $f(s)$ (on average across all feature values for features in $g(s)$ that are not in $f(s)$)
    \item for other states $s' \neq s$, they generate treatment effect values $g(s')^\top \text{proj} \; \Delta_{i,B}, g(s')^\top \text{proj} \; \Delta_{i,N}$ close to the treatment effect values using the original advantage weight vectors $g(s')^\top \Delta_{i,B}, g(s')^\top \Delta_{i,N}$
\end{enumerate}

To find $\text{proj} \; \Delta_{i,B}, \text{proj} \; \Delta_{i,N}$ that achieve both properties, we use the SciPy optimize API\footnote{Documentation here: https://docs.scipy.org/doc/scipy/reference/generated/scipy.optimize.minimize.html} to minimize the following constrained optimization problem:

\begin{align*}
    \min_{\text{proj} \; \Delta} \frac{1}{K} \sum_{k = 1}^K (g(s')_k^\top \text{proj} \; \Delta - g(s')_k^\top \Delta)^2
\end{align*}
\begin{align*}
    \text{subject to:} \;\; \Tilde{g}(s)^\top \text{proj} \; \Delta = 0
\end{align*}

% they satisfy the constraint $\Tilde{g}(s)^\top \text{proj} \; \Delta_{i,B} = 0$ and $\Tilde{g}(s)^\top \text{proj} \; \Delta_{i,N} = 0$.
$\{g(s')_k\}_{k = 1}^K$ denotes a set of states we constructed that represents a grid of values that $g(s')$ could take.
$\Tilde{g}(s)$ has the same state feature values as $g(s)$ except the ``Day of Week" and ``Days Since Participant Started the Trial (Normalized)" features are replaced with fixed mean values $2/7$ and $0$. The objective function is to achieve property 2 and the constraint is to achieve property 1.

We ran the constrained optimization with $\Delta = \Delta_{i, B}$ and $\Delta_{i, N}$  to get $\text{proj} \; \Delta_{i,B}, \text{proj} \; \Delta_{i,N}$, for all participants $i$. All participants in this variant of the simulation environment produce OSCB $Q_{i, t}$ given state $S_{i, t}$ and $A_{i, t}$ using Equation~\ref{eqn_outcome_generating_func} with $\Delta_{i, B}, \Delta_{i, N}$ replaced by $\text{proj} \; \Delta_{i,B}, \text{proj} \; \Delta_{i,N}$.

\section{Additional Did We Learn? Plots}
\label{app_additional_figs}
\begin{figure}[h!]
    \centering
    % first row
    \subfigure[]{
        \includegraphics[width=0.23\textwidth, trim=0cm 0cm 2.5cm 0cm, clip]{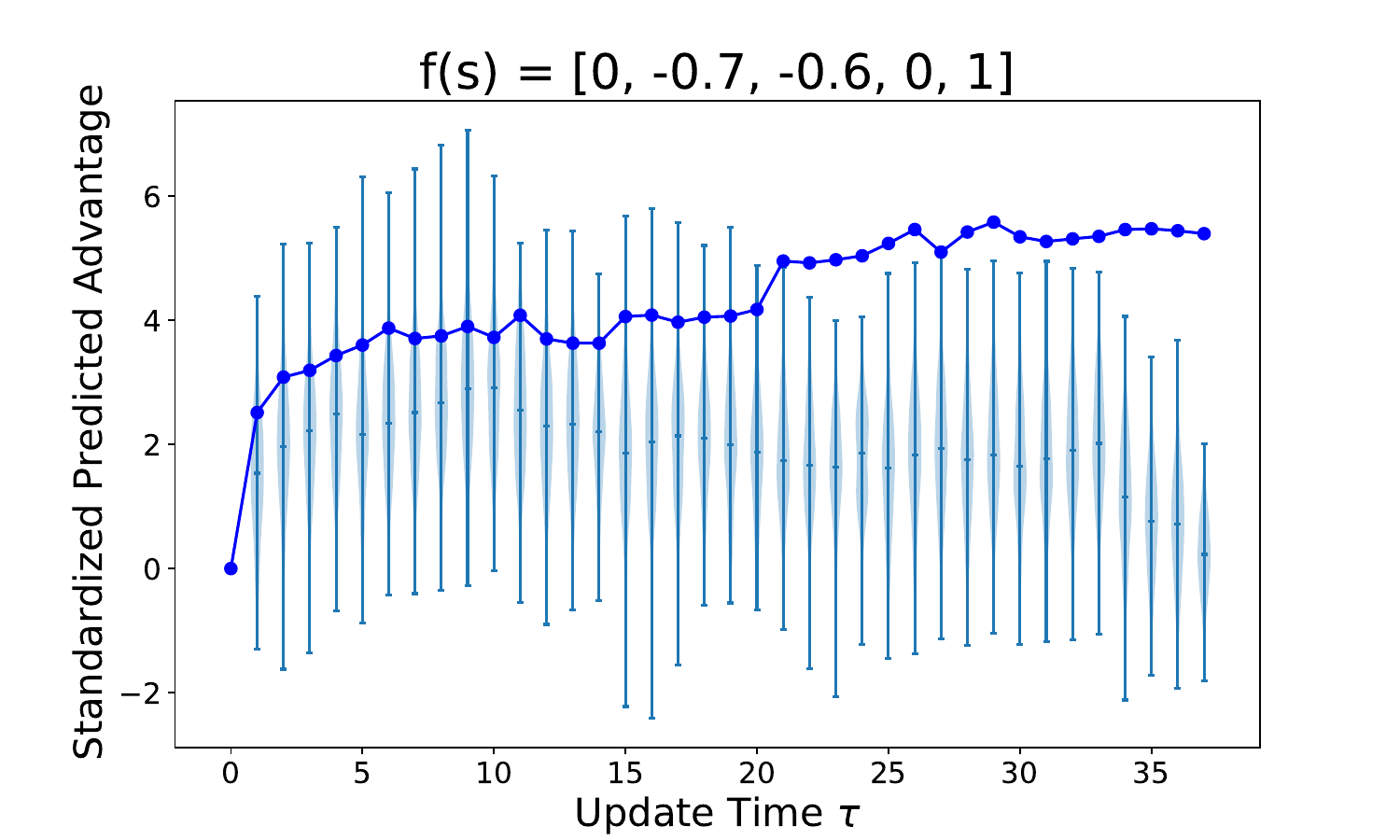}
    }
    \subfigure[]{
        \includegraphics[width=0.23\textwidth, trim=0cm 0cm 2.5cm 0cm, clip]{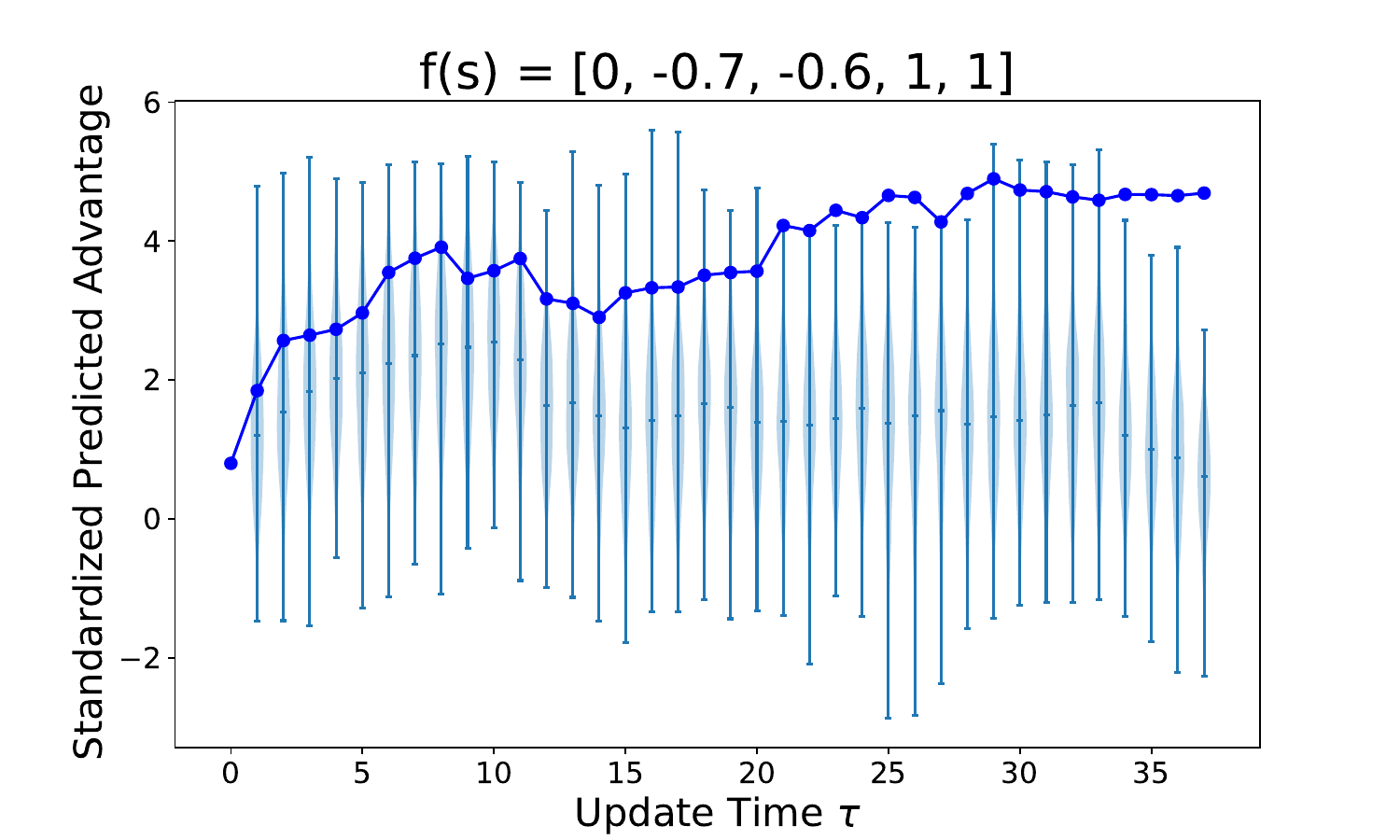}
    }
    \subfigure[]{
        \includegraphics[width=0.23\textwidth, trim=0cm 0cm 2.5cm 0cm, clip]{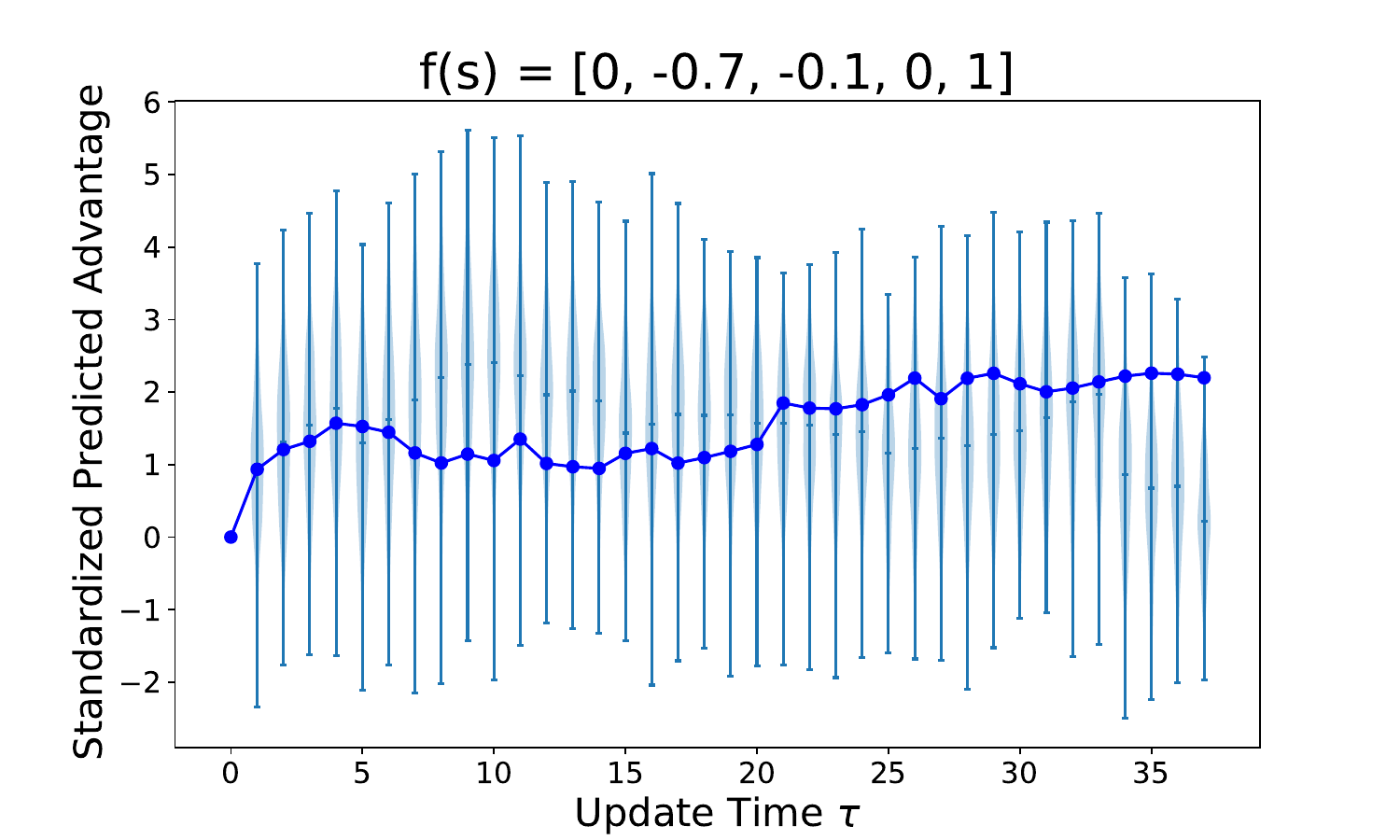}  
    }
    \subfigure[]{
        \includegraphics[width=0.23\textwidth, trim=0cm 0cm 2.5cm 0cm, clip]{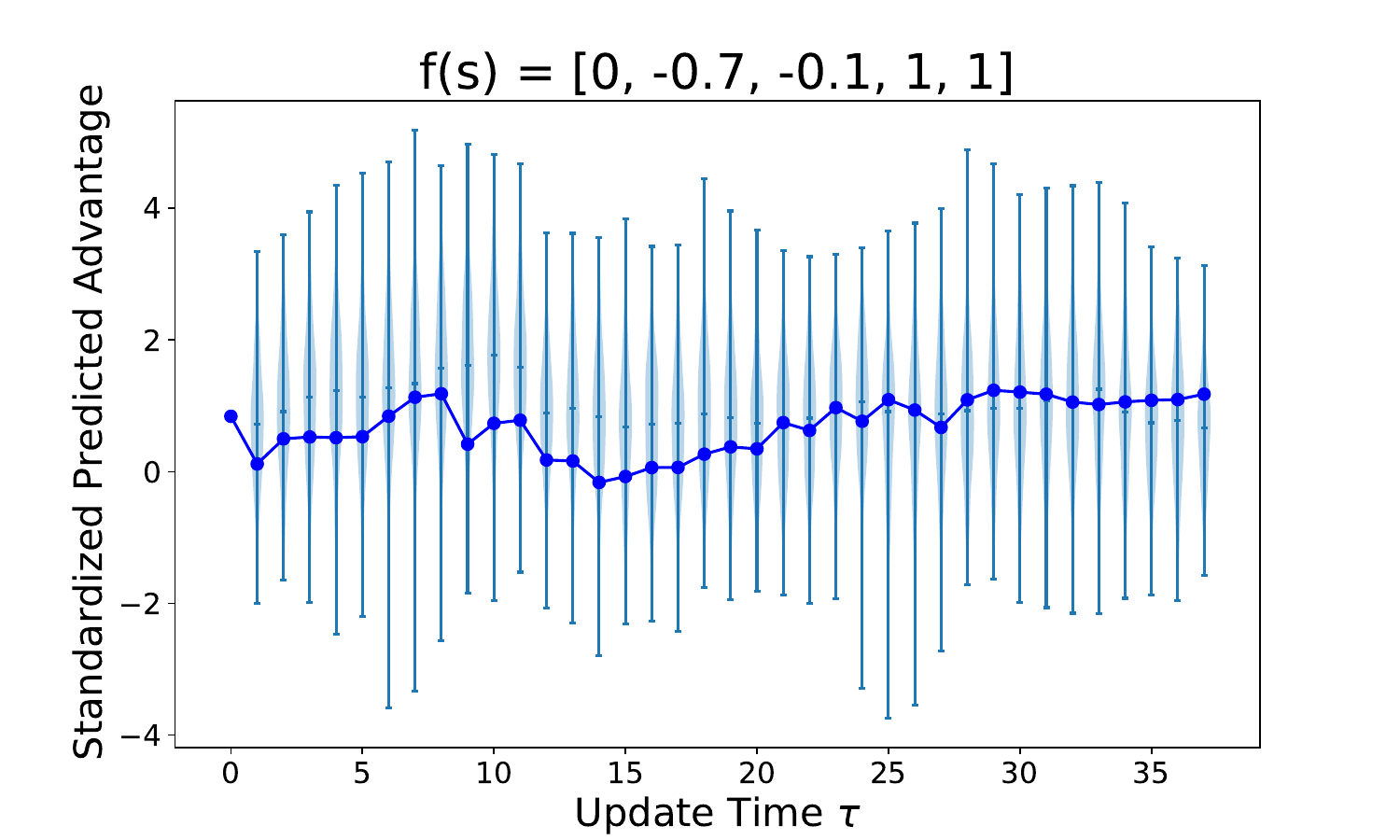} 
    }
    % second row
    \subfigure[]{
        \includegraphics[width=0.23\textwidth, trim=0cm 0cm 2.5cm 0cm, clip]{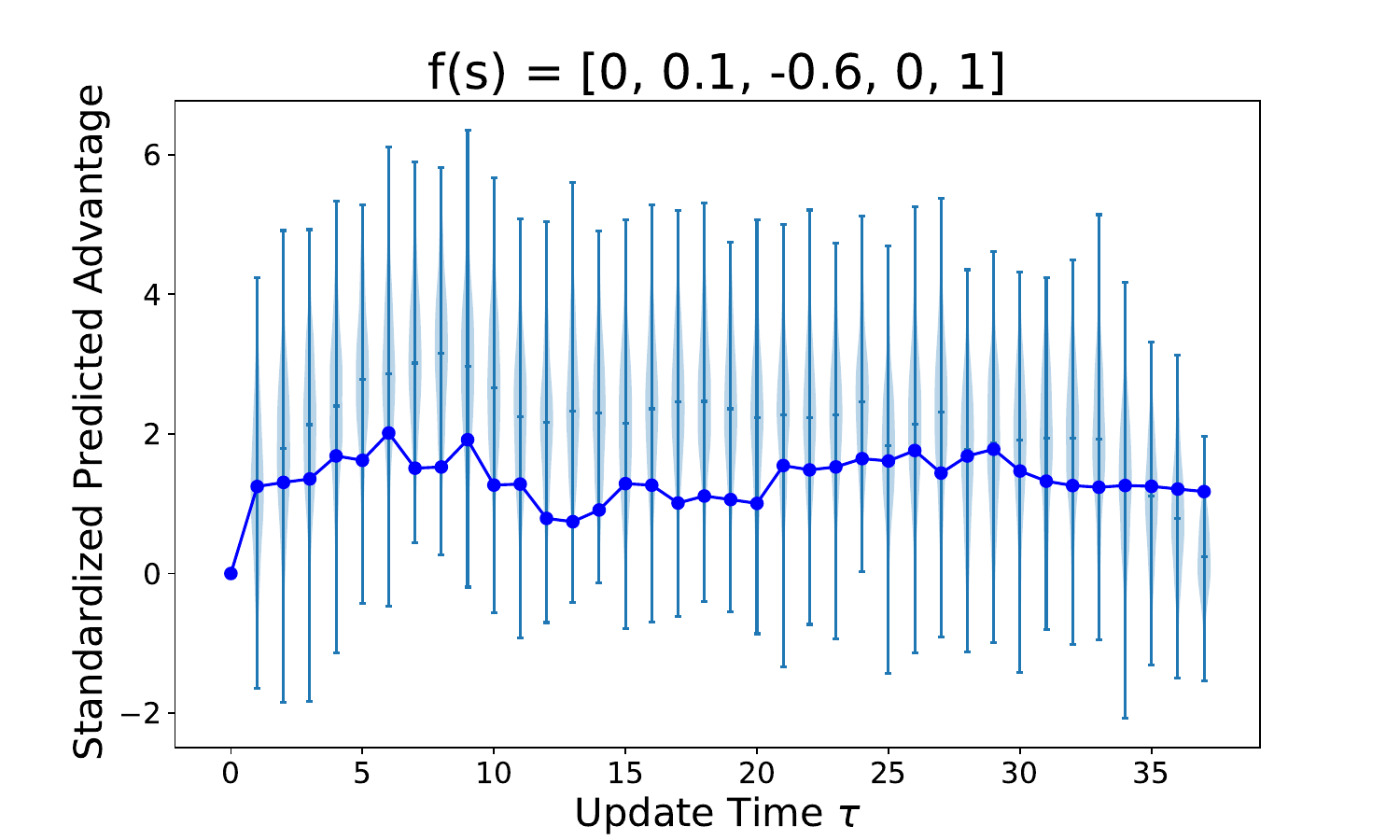} 
        \label{fig:subfig2}
    }
    \subfigure[]{
        \includegraphics[width=0.23\textwidth, trim=0cm 0cm 2.5cm 0cm, clip]{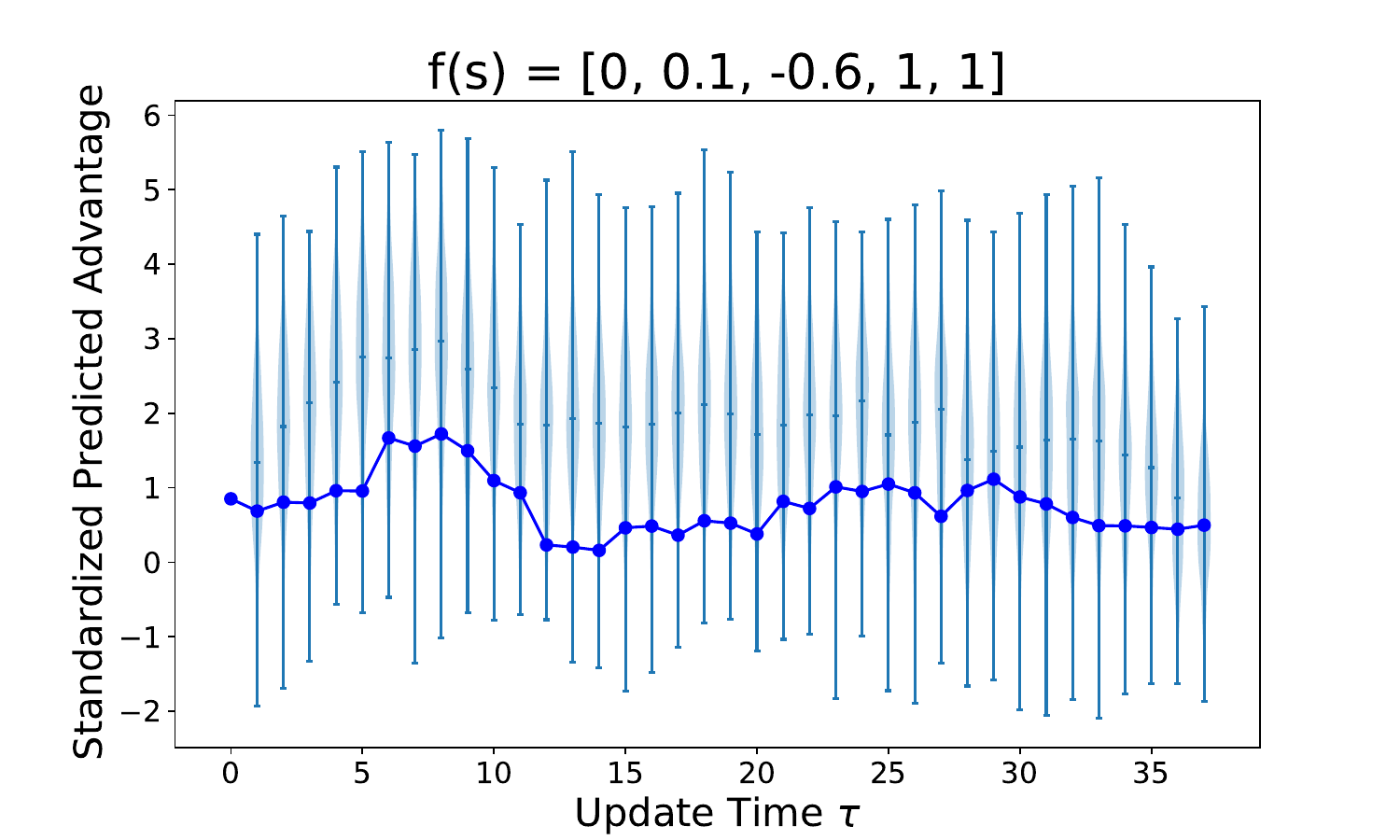}
    }
    \subfigure[]{
        \includegraphics[width=0.23\textwidth, trim=0cm 0cm 2.5cm 0cm, clip]{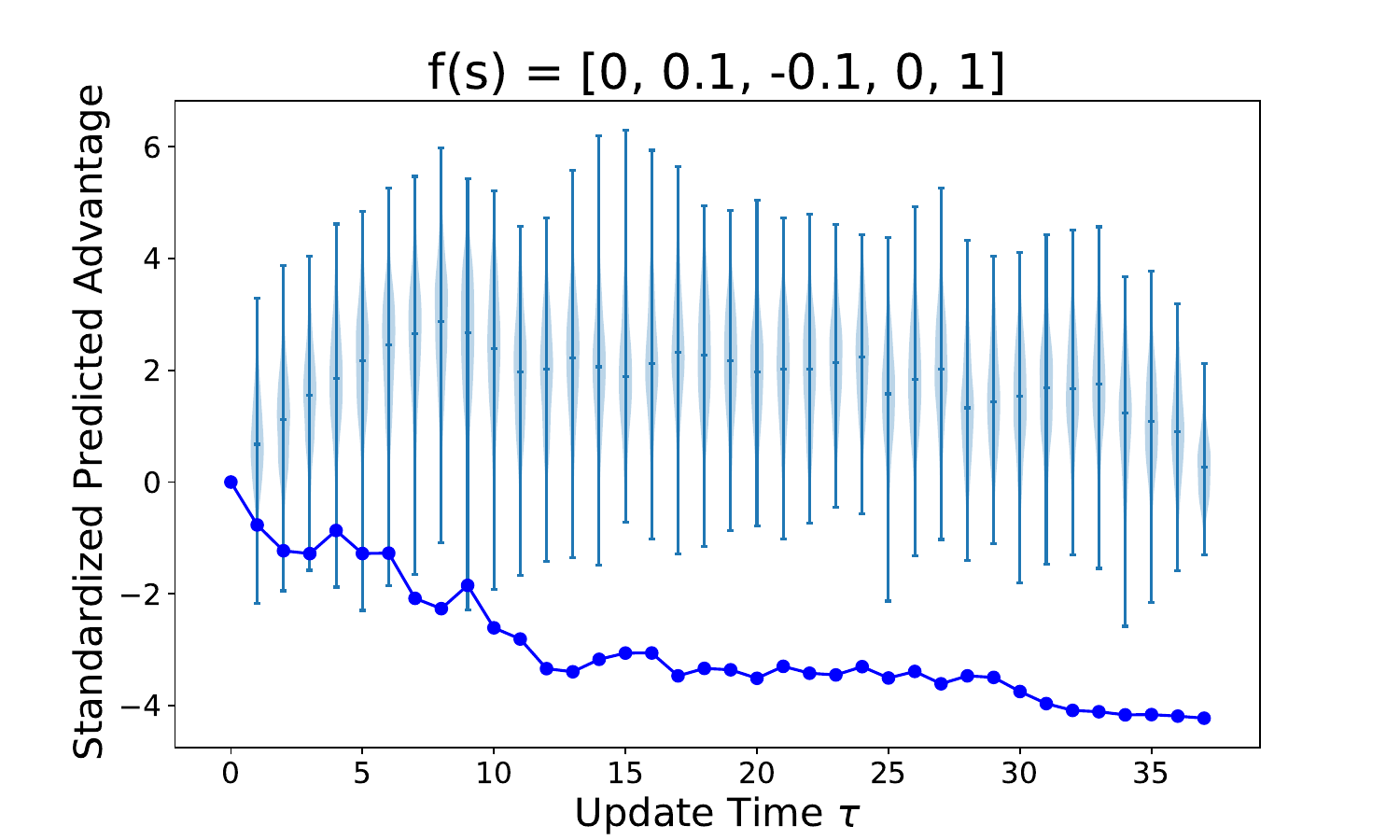}  
    }
    \subfigure[]{
        \includegraphics[width=0.23\textwidth, trim=0cm 0cm 2.5cm 0cm, clip]{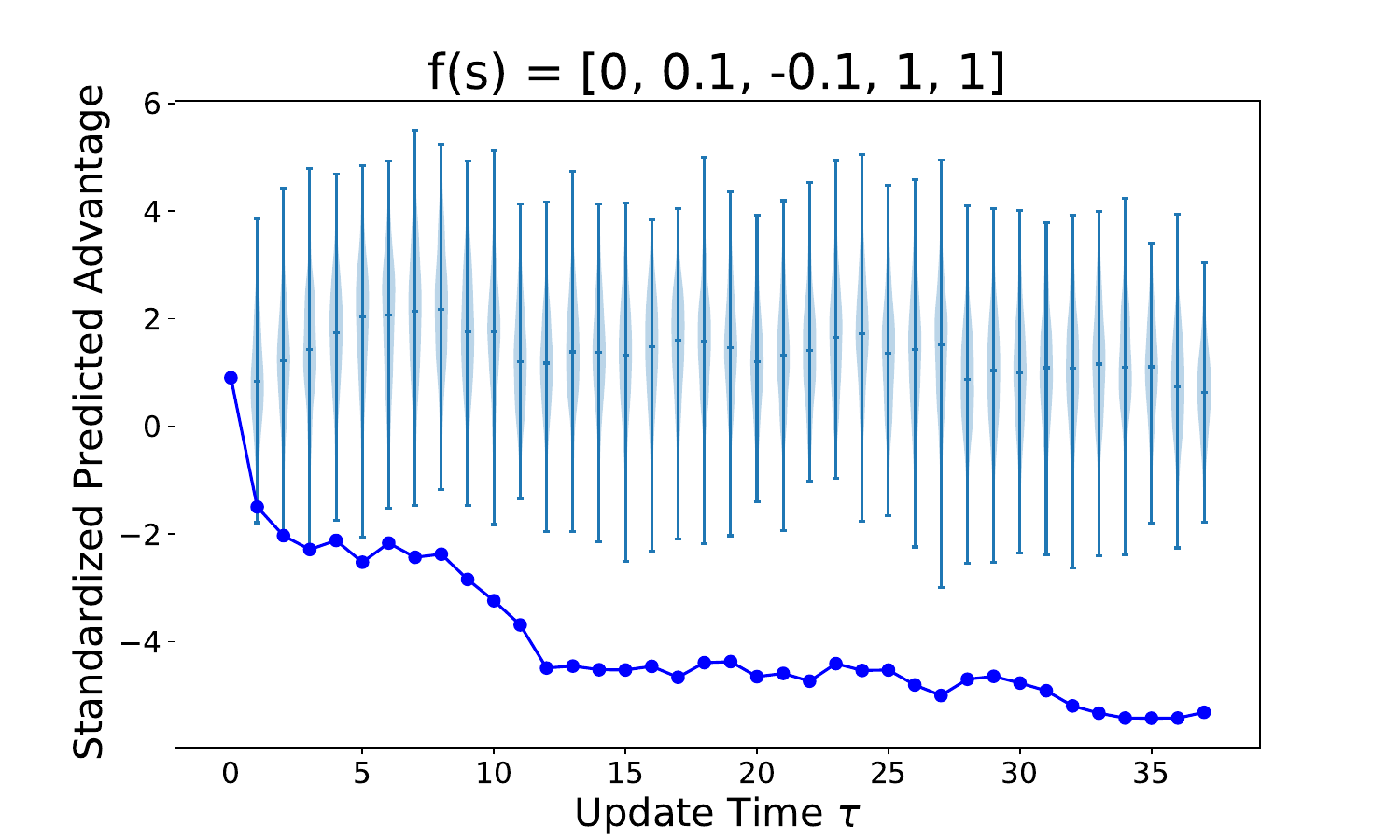} 
    }
    % third row
    \subfigure[]{
        \includegraphics[width=0.23\textwidth, trim=0cm 0cm 2.5cm 0cm, clip]{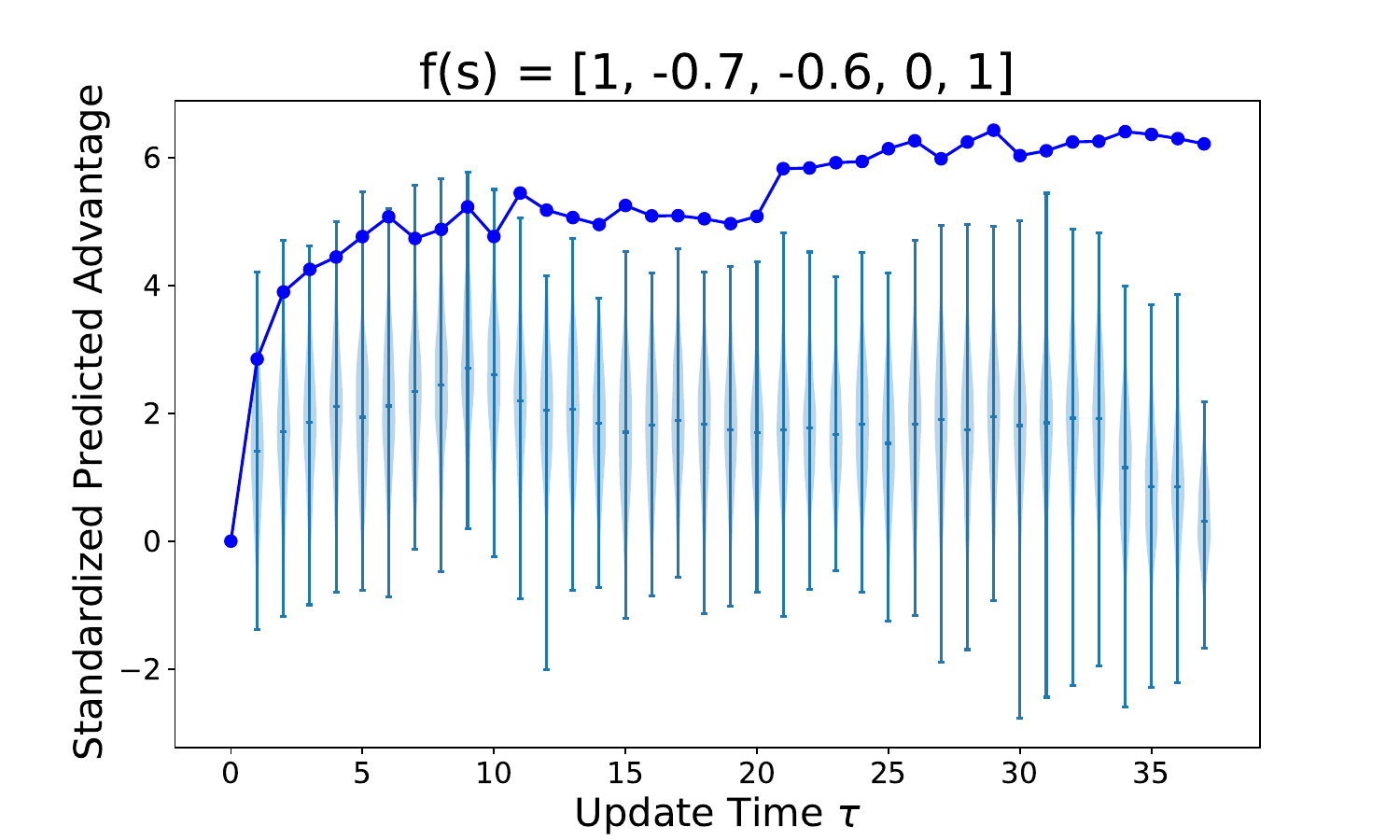} 
    }
    \subfigure[]{
        \includegraphics[width=0.23\textwidth, trim=0cm 0cm 2.5cm 0cm, clip]{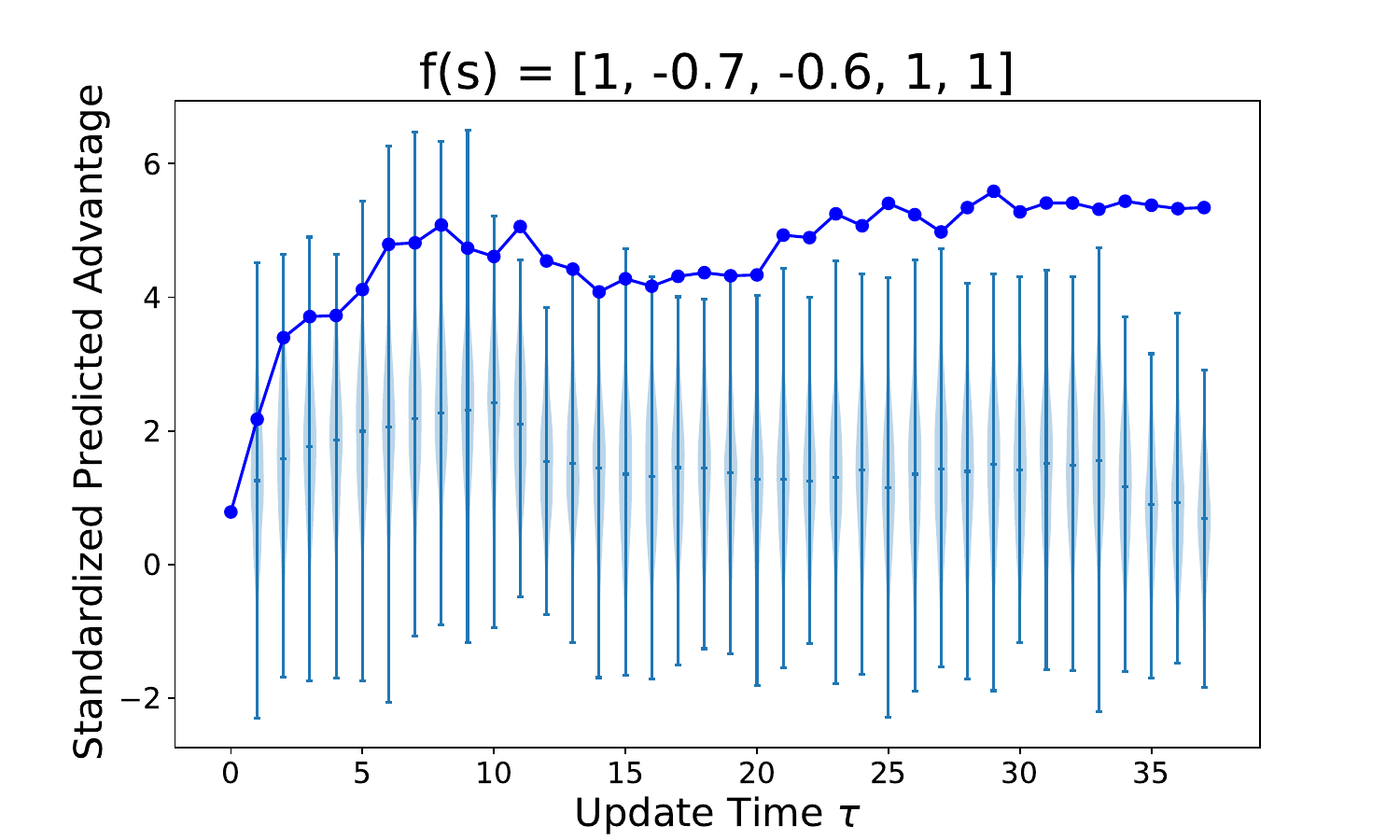}
    }
    \subfigure[]{
        \includegraphics[width=0.23\textwidth, trim=0cm 0cm 2.5cm 0cm, clip]{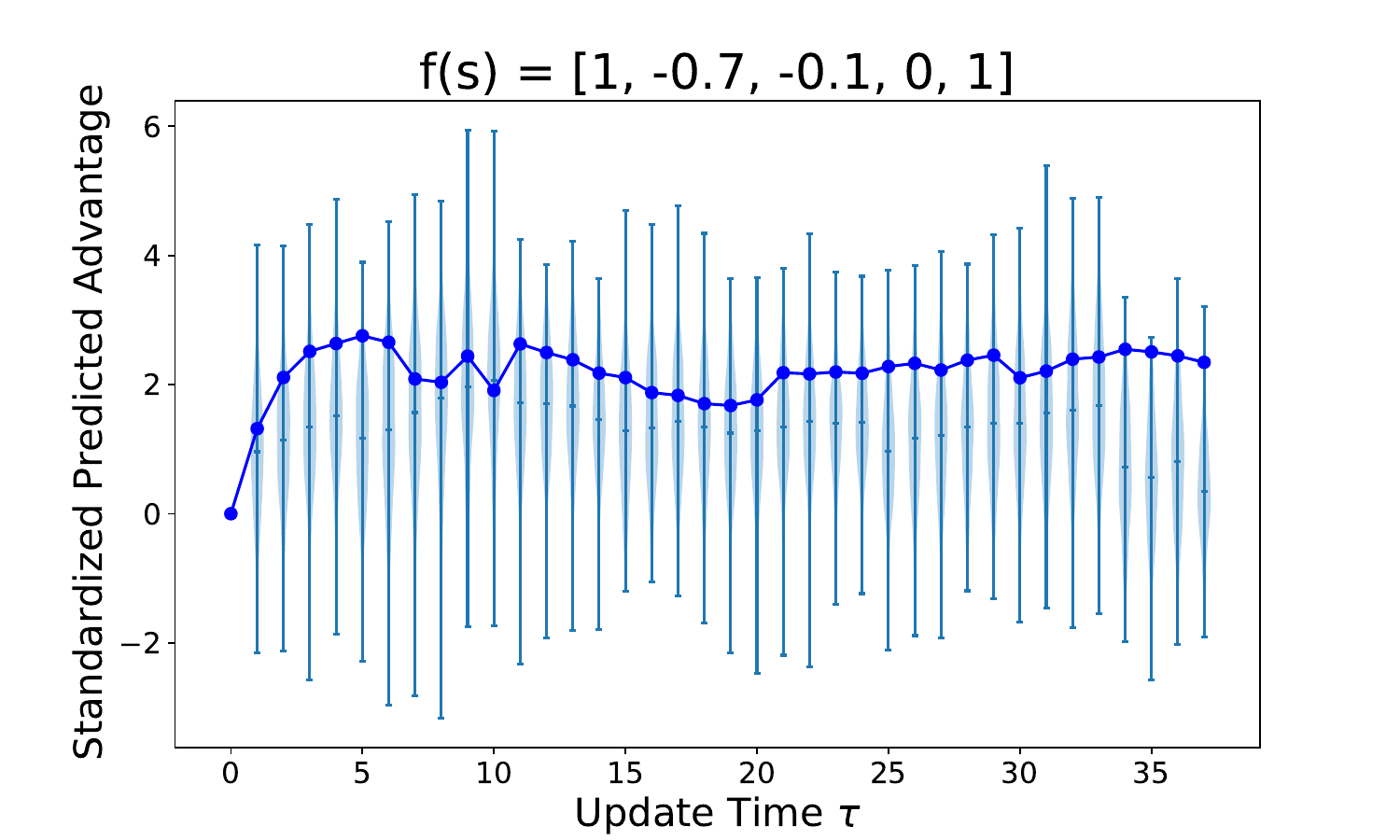} 
    }
    \subfigure[]{
        \includegraphics[width=0.23\textwidth, trim=0cm 0cm 2.5cm 0cm, clip]{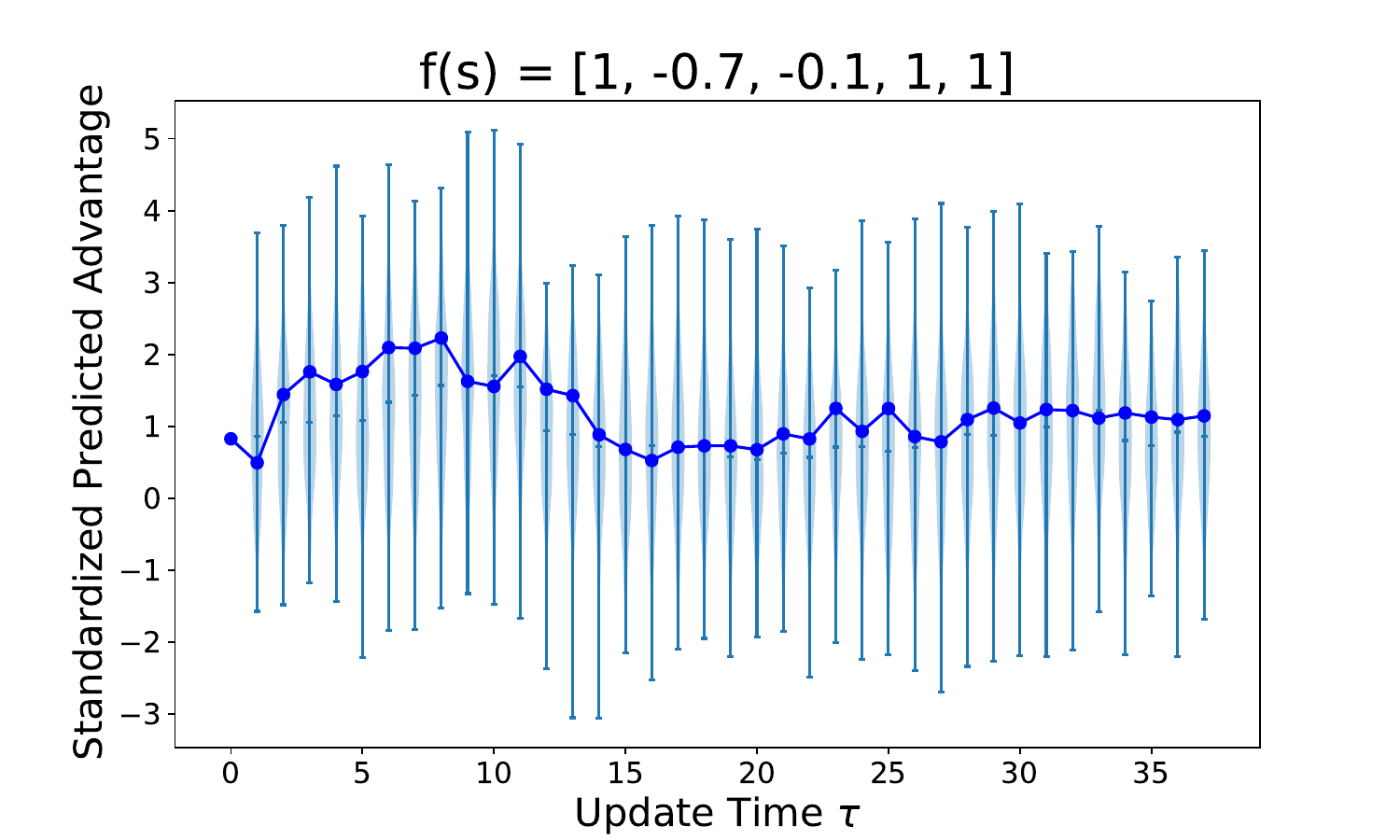} 
    }
    % last row
    \subfigure[]{
        \includegraphics[width=0.23\textwidth, trim=0cm 0cm 2.5cm 0cm, clip]{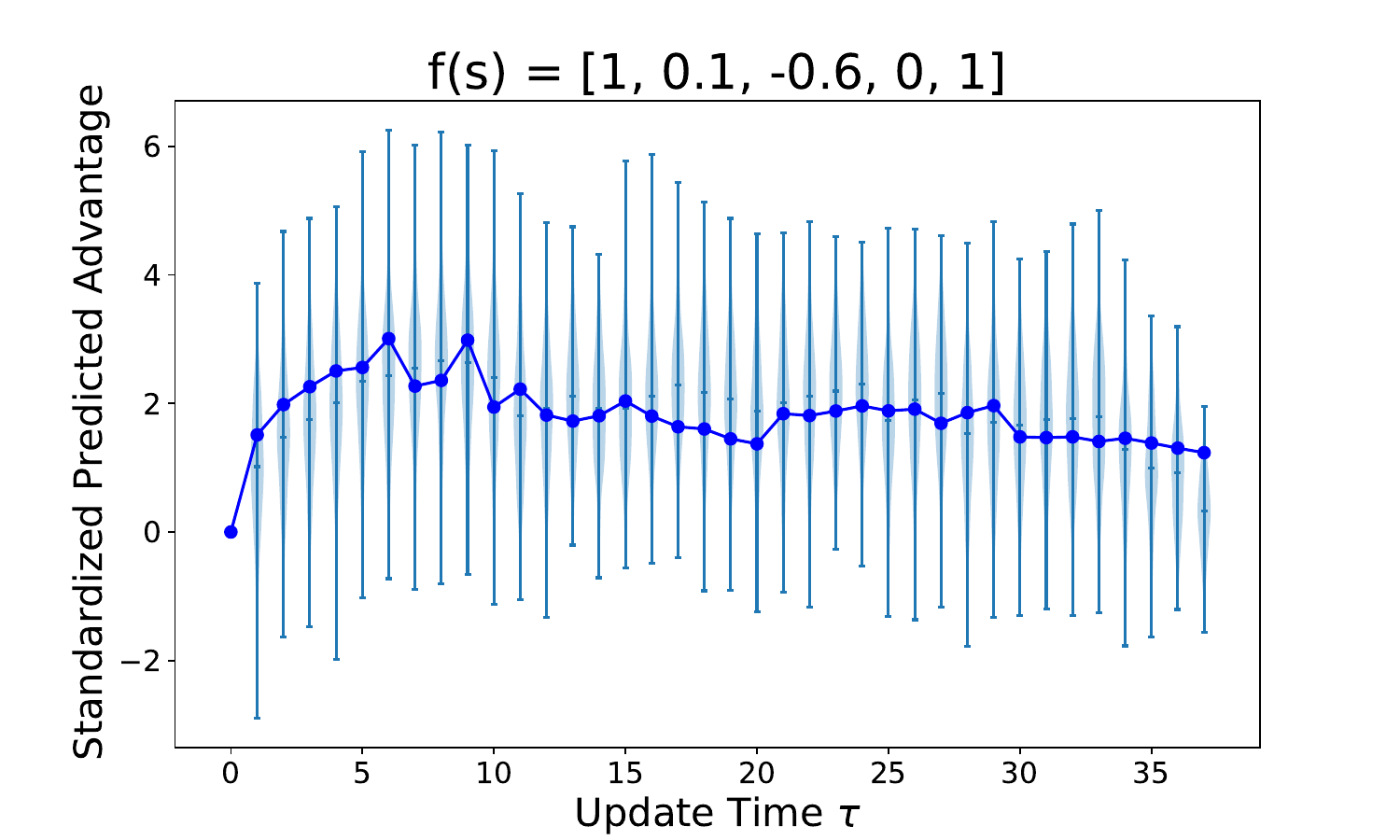} 
    }
    \subfigure[]{
        \includegraphics[width=0.23\textwidth, trim=0cm 0cm 2.5cm 0cm, clip]{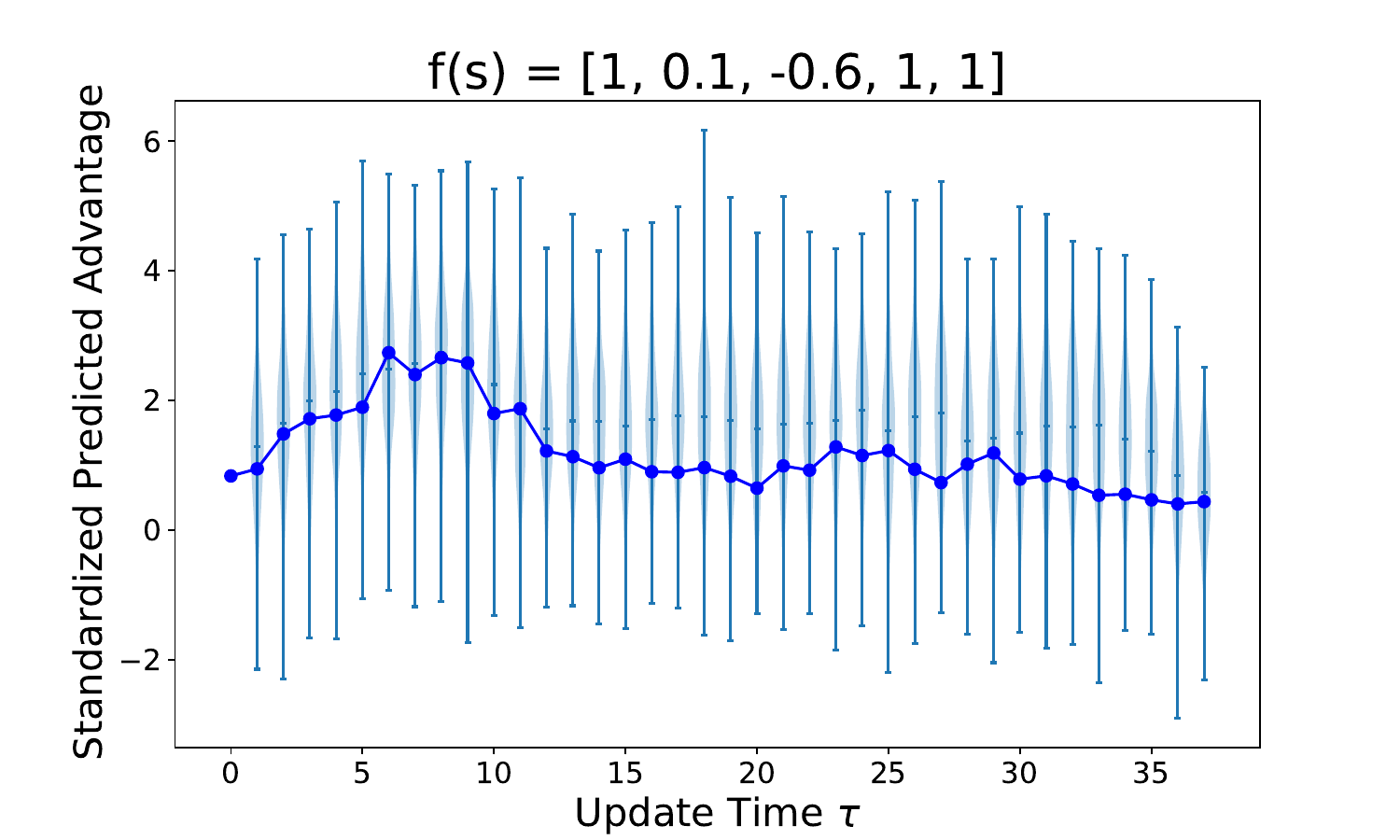}
    }
    \subfigure[]{
        \includegraphics[width=0.23\textwidth, trim=0cm 0cm 2.5cm 0cm, clip]{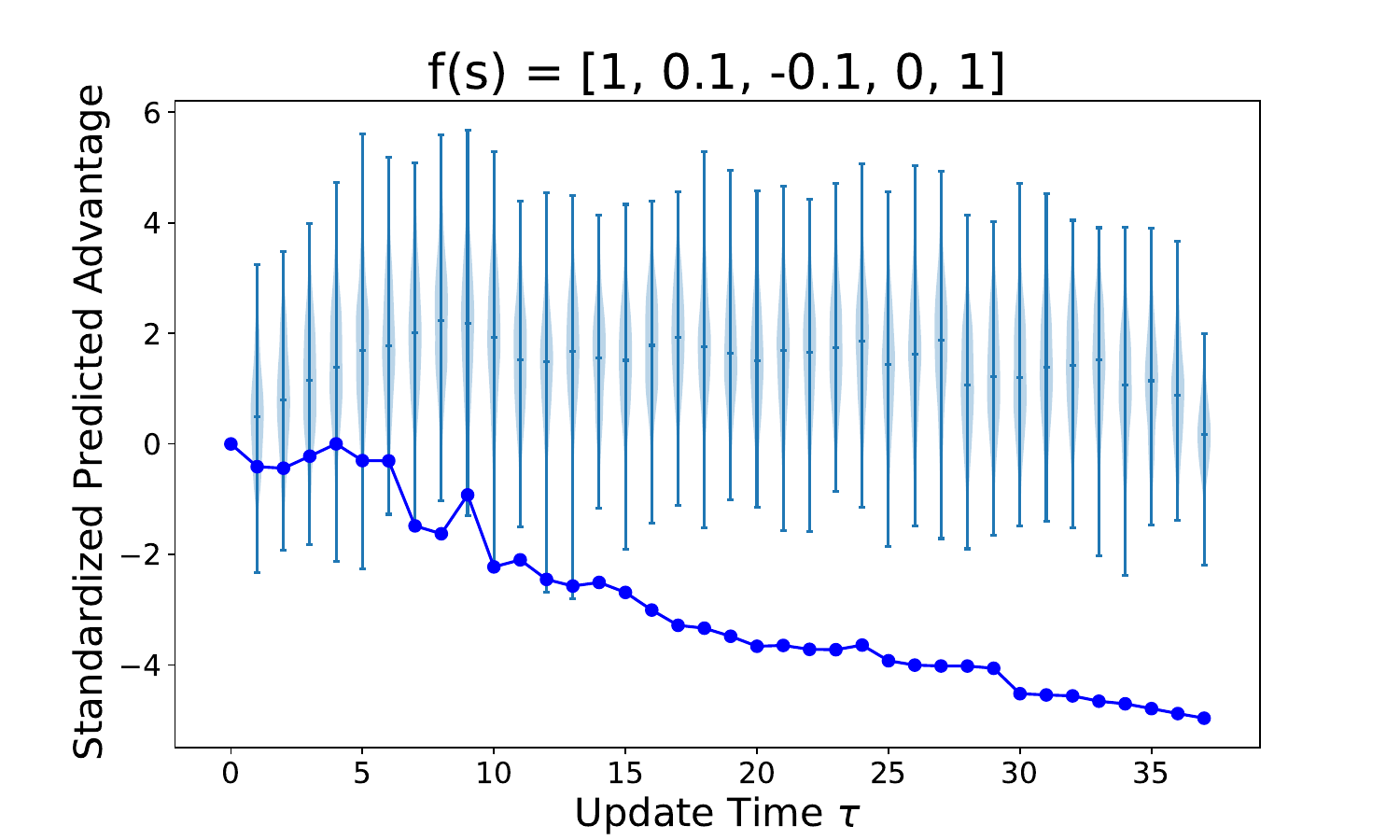} 
    }
    \subfigure[]{
        \includegraphics[width=0.23\textwidth, trim=0cm 0cm 2.5cm 0cm, clip]{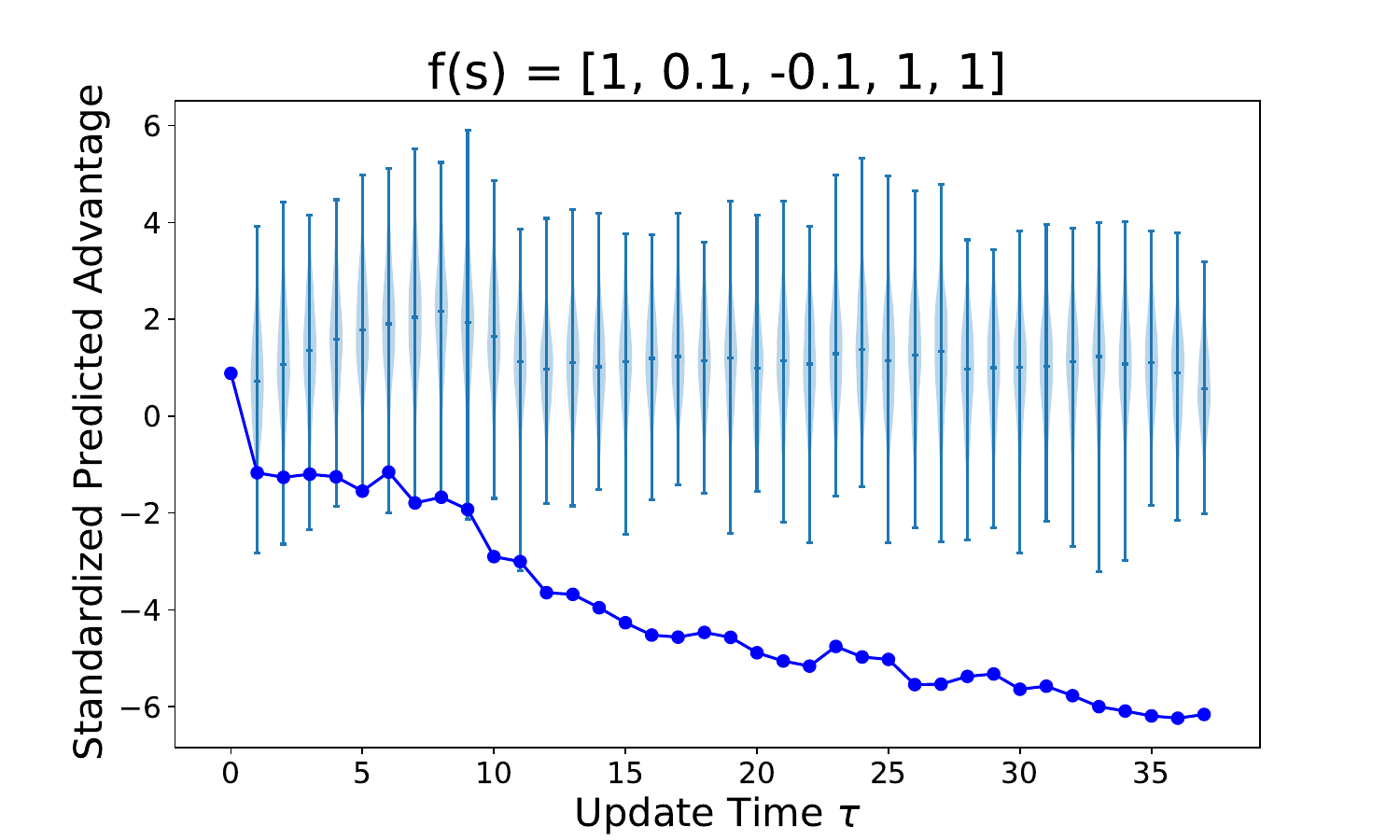} 
    }
    \caption{``Did We Learn?" using the re-sampling based method on 16 different states of interest. We compare standardized predicted advantages across updates to the posterior parameters from the actual Oralytics trial (dark blue) with violin plots of simulated predictive advantages using posterior parameters re-sampled across 500 Monte Carlo repetitions (light blue).}
    \label{fig_did_we_learn_all_16}
\end{figure}

In Section~\ref{did_we_learn} we considered a total of 16 different states of interest. Results for all 16 states are in Figure~\ref{fig_did_we_learn_all_16}. Recall each state is a unique combination of the following algorithm advantage feature values:
\begin{enumerate}
    \item Time of Day: $\{0, 1\}$ (Morning and Evening)
    \item Exponential Average of OSCB Over Past Week (Normalized): $\{-0.7, 0.1\}$ (first and third quartile in Oralytics trial data)
    \item Exponential Average of Prompts Sent Over Past Week (Normalized): $\{-0.6, -0.1\}$ (first and third quartile in Oralytics trial data)
    \item Prior Day App Engagement: $\{0, 1\}$ (Did Not Open App and Opened App)
\end{enumerate}

Notice that since features (2) and (3) are normalized, for feature (2) the quartile value of -0.7 means the participant's exponential average OSCB in the past week is about 28 seconds and similarly 0.1 means its about 100 seconds. For feature (3), the quartile value of -0.6 means the participant received prompts 20\% of the time in the past week and similarly -0.1 means it's 45\% of the time.

\end{document}